\documentclass{egpubl}

\usepackage{subfig}
\usepackage{float}
\usepackage{etoolbox}
\usepackage{xspace}

\SpecialIssuePaper      

\makeatletter
\newcommand\p@copyrightTextShort{}
\newcommand\p@copyrightTextShortEven{}
\newcommand\p@copyrightTextLong{}
\makeatother

\usepackage[T1]{fontenc}
\usepackage{dfadobe}  
\usepackage{amsmath}
\usepackage{bm}
\usepackage{amssymb}
\usepackage{booktabs} 
\usepackage{multirow} 
\usepackage{graphicx}
\biberVersion
\BibtexOrBiblatex
\usepackage[backend=biber,bibstyle=EG,citestyle=alphabetic,backref=true]{biblatex} 
\electronicVersion
\PrintedOrElectronic

\usepackage{egweblnk} 
\newcommand{\ar}[1]{#1}

\newcommand{\projname}{SiGnature\xspace}

\title[SiGnature]
      {SiGnature: Explicit Motion Diffusion \\for Stylized Semantic Gesture Generation}

\author[A. Rosenthal et al.]
{\parbox{\textwidth}{\centering Adi Rosenthal$^{1}$\orcid{0009-0006-6567-696X},
    Tomer Koren$^{1}$\orcid{0009-0000-8472-5727},
    Nadav Shaked$^{1}$\orcid{0009-0002-5615-9478},
    Doron Friedman$^{1}$\orcid{0000-0002-2584-044X},
    and Ariel Shamir$^{1}$\orcid{0000-0001-7082-7845}
    }
    \\
{\parbox{\textwidth}{\centering $^1$Reichman University, Israel
}
}
}

\begin{document}

\teaser{
\includegraphics[width=\textwidth]{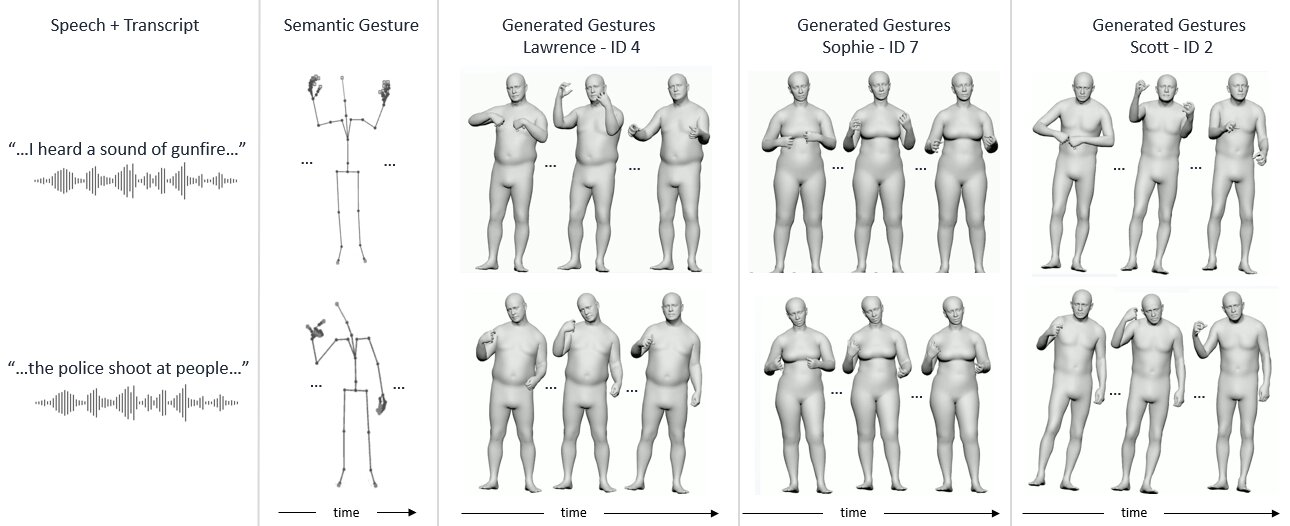}
\centering
\caption{
\textbf{\projname} allows the injection of rare semantic gestures into co-speech motion generation while preserving speaker-specific motion style. For each speech segment (left), we show the corresponding semantic gesture source and the generated gestures for three different identities (Lawrence, Sophie, and Scott).
}
\label{fig:teaser}
}


\makeatletter
\gdef\ps@titlepage{%
  \let\@mkboth\@gobbletwo
  \def\@oddhead{}%
  \def\@evenhead{}%
  \def\@oddfoot{}%
  \def\@evenfoot{}%
  \let\sectionmark\EmptySectionmark
  \let\subsectionmark\EmptySubsectionmark
}
\makeatother

\maketitle
\begin{abstract}
While recent advances in co-speech gesture generation have achieved impressive rhythmic synchronization, synthesizing gestures that are both semantically meaningful and faithful to a speaker's unique non-verbal style remains an open challenge. Semantic gestures, such as iconic shapes or deictic pointing, are statistically sparse, making them difficult to learn effectively within standard generative models. We present \projname, a framework for Stylized and Semantic Gesture generation that reconciles precise semantic control with high-fidelity style preservation.

 Unlike prevalent methods that rely on entangled latent representations, \projname operates in an explicit joint-rotation space. This design enables our core contribution, Joint Motion Integration (JMI), a training-free inference mechanism capable of injecting any external motion sequence, particularly in-the-wild semantic gestures, directly into the diffusion process. JMI automatically identifies the specific ``active joints'' conveying a semantic action and injects them into the generation, while relying on the diffusion backbone to synthesize the remaining body dynamics, including posture and flow, in accordance with the pre-learned style of the target speaker. This allows for the plug-and-play integration of arbitrary motions, including complex semantic gestures, without retraining or introducing the ``Frankenstein'' artifacts typical of cut-and-paste methods. Extensive experiments and perceptual studies demonstrate that \projname offers superior semantic motion control while maintaining smooth and natural co-speech gesture generation and preserving the distinct characteristics of the speaker,  thereby outperforming state-of-the-art baselines.

\end{abstract}  

\section{Introduction}
\label{sec:Introduction}

Human communication is inherently multimodal: hand and body gestures complement speech by conveying emphasis, spatial reference, and semantic nuance~\cite{cassell2000embodied,goldin2013gesture}. For virtual humans to interact naturally in VR/AR and other digital settings, co-speech gestures must be temporally aligned and expressive, while also conveying explicit \emph{semantic} meaning (e.g., iconic, metaphoric, or deictic gestures) and preserving each speaker’s distinctive \emph{signature} style.

Although recent generative models successfully produce beat-synchronized gestures aligned with speech prosody~\cite{zhi2023livelyspeaker,chen2024syntalker,liu2024emage,ijcai2023p650,yi2022generating}, synthesizing semantically meaningful gestures remains a significant challenge. Because these gestures are rare and follow a long-tailed distribution, they are difficult to learn through standard end-to-end training~\cite{cheng2024siggesture,zhang2024semantic}. A key limitation of current approaches is their reliance on compressed latent motion spaces. Although latent manifolds yield strong statistical implicit feature distributions, they lack joint-level granularity and frequently suffer from high-frequency kinematic jitter. Furthermore, these spaces are not directly editable. Even modular ``separate-and-combine'' designs~\cite{chen2024syntalker,liu2024emage} focus on broad body regions rather than specific joints, making local edits prone to degraded overall style or motion coordination. These limitations become critical when attempting to inject external, rare semantic motions. Because long-tailed gestures are typically out-of-distribution for standard VAEs, they suffer from severe reconstruction issues. In VQ-VAE codebooks specifically, this leads to quantization artifacts that ``snap'' the generated motions to generic patterns, erasing the intended semantic meaning.

To address these limitations, we propose \projname, a two-part framework designed to reconcile precise semantic control with high-fidelity style preservation. First, we introduce a multimodal diffusion backbone operating directly in \textbf{explicit rotation space}, enabling precise joint-level control. Second, we present \textbf{Joint Motion Integration (JMI)}, a training-free inference mechanism that uses a spatial variance heuristic to automatically isolate and inject only the active semantic joints from an external motion clip. Simultaneously, our personalized diffusion prior actively synthesizes the remaining body dynamics. This effectively decouples the injected semantic action from the target speaker's unique dynamic style, ensuring coherent, localized generation without sacrificing global style or creating cut-and-paste artifacts.

We evaluate \projname backbone against state-of-the-art co-speech generative baselines, demonstrating that moving to an explicit-space architecture does not degrade general co-speech synthesis. In contrast, it achieves what we term \emph{``active smoothness''}, a combination of high expressive diversity and kinematic fluidity (low Jerk/Accel). We then evaluate our full pipeline (featuring the JMI mechanism) against existing semantic injection methods. Quantitative metrics and perceptual user studies confirm that \projname produces highly natural, human-like co-speech motion that is seamlessly editable, while preserving the target speaker's style.

Our main contributions are as follows:

\vspace{-0.5em}

\begin{itemize}
\ar{\item \textbf{Co-Speech Diffusion in Joint-Rotation Space}: We trained a multimodal co-speech motion generator that diffuses directly in an explicit joint-rotation space, rather than a compressed latent space. This approach improves kinematic smoothness and diversity while enabling precise per-joint editing.}


\item \textbf{Joint Motion Integration (JMI)}: A training-free mechanism that uses spatial variance analysis to \ar{ identify active joints, then integrates arbitrary semantic gestures via per-joint in-betweening,} effectively addressing the long-tailed distribution problem of semantic actions.

\item \textbf{Style-Preserving Semantic Synthesis}: Our framework successfully decouples semantic action from speaker style, allowing for the generation of explicit semantic gestures that naturally coexist with the unique dynamics of a specific target speaker.

\end{itemize}

We will make our source code, datasets, and pre-trained models publicly available upon publication.

\begin{figure*}[!t]
  \centering
  \includegraphics[width=0.9\textwidth, height=0.4\textwidth]{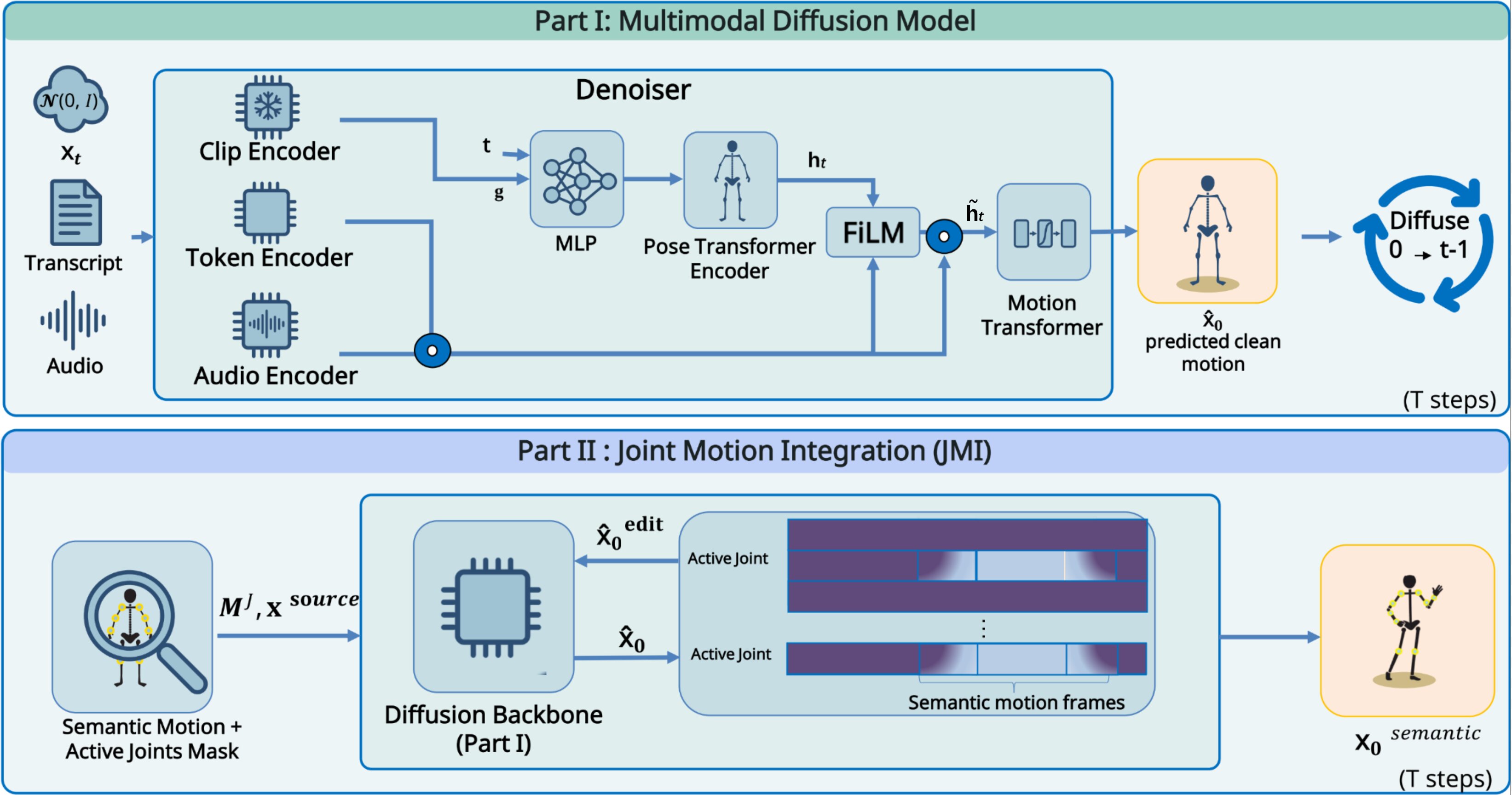}
  \vspace{-0.2cm}
  \caption{\textbf{Overview of our two-part architecture.}
\textbf{Part I - Multimodal diffusion model.}
Starting from Gaussian noise $\mathbf{x}_T \sim \mathcal{N}(\mathbf{0},\mathbf{I})$,
the model iteratively denoises in SMPL-X rotation space.
At each step $t$, the noisy motion sequence $\mathbf{x}_t$ is embedded as motion tokens
and fused with a timestep embedding summed with a global sentence-level CLIP embedding.
Both transcript tokens and audio are encoded. The resulting frame-aligned audio--text features
are passed through a FiLM network {\cite{perez2018film}} to modulate the motion embeddings,
and are also \emph{concatenated} with them before being processed by the motion transformer.
This multimodal diffusion backbone produces a clean motion prediction $\hat{\mathbf{x}}_0$.
\textbf{Part II - Joint Motion Integration (JMI).}
During inference, the trained diffusion backbone is used and augmented with semantic gesture
motion clips.
The \emph{Find Active Joints} module computes active-joint masks $M^J$ for each
semantic gesture motion.
JMI injects these semantic joint trajectories only in the selected active joints
and frames of the backbone prediction, yielding an edited sequence
$\hat{\mathbf{x}}_0^{\text{edit}}$ and a final motion
$\hat{\mathbf{x}}_0^{\text{semantic}}$ that preserves the speaker’s style while
enforcing the desired semantic gestures across diffusion steps.}
\label{fig:concept}
\vspace{-5mm}
  \label{fig:architecture}
\end{figure*}

\section{Related Work}

\textbf{Text-to-Motion Generation.}
Text-driven human motion synthesis has evolved from sequence-to-sequence models to learning shared latent spaces with vision-language models and diffusion priors.
\textit{MotionCLIP}~\cite{tevet2022motionclip} aligns a motion auto-encoder with CLIP’s text-image space, enabling text-driven generation and editing.
Following this, the \textit{Motion Diffusion Model} (MDM) ~\cite{tevet2023human} utilizes a transformer-based architecture that predicts the clean motion directly rather than the noise residual. This formulation enables the use of geometric losses (e.g., foot contact) and supports ``in-betweening''-style editing by constraining specific frames or joints during the sampling process.
Building on this, PriorMDM ~\cite{shafir2024human}, with its \textit{Double-Take} inference scheme, enables long-form generation by overlapping prompted intervals. 
\ar{Although our architecture shares the clean-signal prediction formulation of MDM \cite{tevet2023human} and the Double-Take inference strategy~\cite{shafir2024human}, we adapt it from the text-only setting to the multimodal co-speech domain. A key distinction is the dual-stream conditioning within our diffusion model, which enables the rhythmic and stylistic alignment that text-to-motion backbones do not natively support.}

\textbf{Co-Speech Gesture Generation.}
Early approaches aligned motion with prosodic cues using RNNs or transformers~\cite{yoon2019robots,ginosar2019learning}, often resulting in repetitive beat gestures with little semantic meaning.
Subsequent methods scaled up training data and modalities~\cite{ahuja2019language2pose,yoon2020speech,li2021audio2gestures,liu2022learning,yi2022generating,chen2024body_of_language}, with some approaches adopting VQ-VAEs to map speech and transcript to full-body motion. \textit{EMAGE}~\cite{liu2024emage} further advanced this by employing part-wise VQ-VAEs (face, hands, body) and a masked gesture transformer to achieve holistic control.

Recent works leverage high-capacity generative models~\cite{zhu2023taming,xu2024mambatalk,chen2024diffsheg} to improve audio-motion correlation.
\textit{GestureLSM}~\cite{liu2025gesturelsm} introduces a flow-matching framework with spatial-temporal attention, aiming to resolve the disjointed movements of part-based models and accelerate sampling speeds. \textit{SynTalker}~\cite{chen2024syntalker} augments co-speech datasets with text-to-motion data and introduces a diffusion-based, multi-stage pipeline with a \emph{separate-then-combine} inference strategy, enabling prompt-based control. However, while this allows for high-level guidance, it lacks the precision of direct motion injection for specific semantic gestures.
In parallel, several diffusion-based co-speech methods emphasize speaker style or controllability~\cite{ijcai2023p650,Ao2023GestureDiffuCLIP,qi2024emotiongesture,Chhatre_2024_CVPR,EMOTE,ghorbani2022zeroeggs}.
Similar to \textit{GestureDiffuCLIP}~\cite{Ao2023GestureDiffuCLIP}, which utilizes CLIP text embeddings to learn a joint gesture-transcript embedding space, we also leverage CLIP-derived transcript features. However, rather than using them for latent space alignment, we inject these features as a global semantic prior into our diffusion backbone.

\textbf{Semantic Gesture Integration and Personalization.}
To address the scarcity of semantically meaningful gestures in standard datasets, recent works attempt to decouple base prosody from semantic intent. For instance, \textit{SemTalk}~\cite{zhang2025semtalk} learns rhythm-related base motions and sparse semantic motions in separate latent spaces, adaptively fusing them via a learned semantic score. Other hybrid retrieval-generation approaches have been proposed. \textit{Semantic Gesticulator}~\cite{zhang2024semantic} uses an LLM to detect gesture-relevant transcript segments, retrieves a labeled gesture clip, and blends it in latent motion space. \textit{SIGGesture}~\cite{cheng2024siggesture} similarly performs latent-space fusion by injecting a semantic representation into the diffusion trajectory, merging semantic motion with rhythmic co-speech dynamics. More recently, \textit{RAG-Gesture} ~\cite{mughal2025retrieving} implies inference-time injection via DDIM inversion, modifying the noise space to guide generation toward a retrieved exemplar. While effective, such continuous latent-space or codebook-based operations rely on pretrained embeddings that may override speaker-specific style or degrade fidelity; adding new complex gestures often requires costly retraining or codebook extension.

\textbf{Limitations of Current Approaches.}
A critical limitation of state-of-the-art approaches is their heavy reliance on compressed latent spaces. Because these representations encode motion holistically, they inherently lack joint-level control. This makes it exceedingly difficult to perform localized edits without unintentionally altering other regions and degrading the speaker’s personal gesture style. Furthermore, when rare, long-tailed semantic gestures (e.g., intricate hand articulations) are projected into a latent space, the compression process frequently fails to preserve fine-grained details. Discrete models (e.g., VQ-VAEs) risk ``snapping'' distinct, out-of-distribution motions to generic codebook priors, while continuous spaces tend to over-smooth high-frequency kinematics due to manifold projection. In both cases, the intended semantic meaning is effectively erased or severely distorted during reconstruction.

\textbf{Our Approach.}
In contrast to previous methods, we introduce an editable co-speech generator that performs \emph{semantic injection directly in SMPL-X rotation space}. By avoiding latent compression, our method enables fine-grained editing of co-speech gestures while preserving high-fidelity motion details and speaker-specific gestural characteristics. Through our Joint Motion Integration (JMI) module, we identify and modify only the active joints in the injected motion, enabling smooth, training-free integration of arbitrary, in-the-wild semantic gestures without the ``Frankenstein'' artifacts or style degradation often associated with cut-and-paste or latent-blending methods.
\section{Method}
Our goal is to synthesize natural and expressive co-speech gestures that are aligned with both the audio beat and the semantic content of the utterance, while preserving the target speaker's unique style. To achieve this, our proposed framework consists of two main components (see Fig.~\ref{fig:architecture}): \textbf{Part I}, a trained multimodal diffusion backbone that operates directly in rotation space; and \textbf{Part II}, Joint Motion Integration (JMI), a training-free inference editor that inserts meaningful gestures in joint space while preserving the speaker’s style.

\subsection{Problem Setup and Representation}

Let a motion clip consist of per-frame joint rotations, root translation, and foot-contact indicators. For each frame $n$, let $\mathbf{r}_n \in \mathbb{R}^{J \times 6}$ denote the 6D continuous local joint rotations for $J$ joints, $\mathbf{p}_n \in \mathbb{R}^3$ the global root translation, and $\mathbf{f}_n \in \{0,1\}^{4}$ the binary foot-contact flags. We flatten these into a motion feature vector per-frame:
\[
\mathbf{x}_n = \bigl[\operatorname{vec}(\mathbf{r}_n),\ \mathbf{p}_n,\ \mathbf{f}_n\bigr] \in \mathbb{R}^{C},
\]
where $C = 6J + 3 + 4$ is the total feature dimension. The full motion sequence is denoted as $\mathbf{x} \in \mathbb{R}^{N \times C}$, where $N$ is the number of frames in the motion window.

We tokenize the utterance transcript into a frame-level sequence $\mathbf{s}_{1:N}$, where $\mathbf{s}_n$ is the text token active at frame $n$. Concurrently, we extract the corresponding raw audio waveform $\mathbf{a}_{1:N_a}$. The length of the raw audio sequence, $N_a$, is determined by the frame rate and audio sample rate. We additionally compute a global sentence-level transcript embedding $\mathbf{g} \in \mathbb{R}^d$ using a frozen CLIP text encoder. Unlike the frame-level tokens, this embedding provides a global semantic prior (see Sec.~\ref{par:cond}). 

\subsection{Part I: Multimodal Diffusion Backbone}
\label{sec:Multimodal-Diffusion-Backbone}
We formulate co-speech gesture generation as a conditional denoising diffusion process in explicit motion space. Let $\mathbf{x}_0 \in \mathbb{R}^{N \times C}$ denote the ground-truth motion sequence. The forward process gradually perturbs $\mathbf{x}_0$ with Gaussian noise,
\[
q(\mathbf{x}_t \mid \mathbf{x}_{t-1})
= \mathcal{N}\!\bigl(\sqrt{1-\beta_t}\,\mathbf{x}_{t-1},\, \beta_t \mathbf{I}\bigr),
\quad t = 1,\dots,T
\]
using a variance schedule $\{\beta_t\}_{t=1}^T$. During inference, we start from $\mathbf{x}_T \sim \mathcal{N}(\mathbf{0},\mathbf{I})$ and iteratively denoise toward a clean motion sample.

\textbf{Denoising objective.}
We employ a transformer-based denoiser $f_\theta$ that predicts the clean motion $\hat{\mathbf{x}}_0$ from a noisy input $\mathbf{x}_t$ at diffusion step $t$:
\[
\hat{\mathbf{x}}_0
= f_\theta\!\bigl(
\mathbf{x}_t,\;
t,\;
\mathbf{a}_{1:N_a},\;
\mathbf{s}_{1:N},\;
\mathbf{g}
\bigr).
\]

The primary reconstruction objective is an $\ell_2$ loss in rotation space, $\mathcal{L}_{\text{rec}} = \sum_{k} \|\mathbf{r}_k - \hat{\mathbf{r}}_k\|_2^2$. To improve temporal coherence, we additionally penalize rotation-space velocities, $\mathcal{L}_{\text{vel}} = \sum_{k} \|\mathbf{v}_k - \hat{\mathbf{v}}_k\|_2^2$, where $\mathbf{v}_k = \mathbf{r}_k - \mathbf{r}_{k-1}$. Finally, we enforce physical plausibility via a foot contact loss $\mathcal{L}_{\text{foot}}$ that penalizes foot sliding when the ground-truth contact mask is active. The total objective is:
\[
\mathcal{L}
= \mathcal{L}_{\text{rec}}
+ \lambda_{\text{vel}}\,\mathcal{L}_{\text{vel}}
+ \lambda_{\text{fc}}\,\mathcal{L}_{\text{foot}}.
\]

\textbf{Conditioning and fusion.}
\label{par:cond}
We decouple the conditioning signals into two parallel streams to effectively integrate global semantic style with local rhythmic synchronization:

\noindent\textbf{1. Global Semantic Prior.}
To inject the global semantic intent $\mathbf{g}$, we project it and add it directly to the diffusion timestep embedding $\mathbf{t}_{\text{emb}}$. To incorporate this into the transformer architecture, the fused global embedding is appended to the input motion sequence along the temporal dimension (acting as a global context token). By fusing $\mathbf{g}$ into the temporal backbone, we bias the entire denoising trajectory toward the high-level semantic context, independent of the local audio prosody.

\noindent\textbf{2. Local Prosodic Style Modulation.}
To improve precise stylistic articulation and frame-level alignment with speech, we process the raw audio waveform $\mathbf{a}_{1:N_a}$ and transcript tokens $\mathbf{s}_{1:N}$ through lightweight encoders (1D convolutions and linear embeddings, respectively) and concatenate them along the feature dimension to form a localized condition vector $\mathbf{c} = [\mathcal{E}_a(\mathbf{a}) \,;\, \mathcal{E}_s(\mathbf{s})]$. This local condition sequence serves a dual purpose within our architecture: First, it drives a Feature-wise Linear Modulation (FiLM)~\cite{perez2018film} network to predict dense, time-varying  scale and shift parameters $\bm{\Gamma}, \mathbf{B} \in \mathbb{R}^{T \times D}$, matching the temporal and feature dimensions of the intermediate hidden features. These parameters apply a fine-grained, element-wise affine transformation to the hidden motion features $\mathbf{h}_t$ at every frame $t$:
\[
    \mathbf{h}'_{t} = \mathbf{h}_{t} \odot (1 + \bm{\Gamma}_{t}) + \mathbf{B}_{t},
\]

where $\mathbf{h}_t$ represents the intermediate motion features (after processing the global semantic prior) and $\mathbf{h}'_t$ represents the resulting modulated motion features. This mechanism allows the audio energy and word timing to dynamically amplify or suppress specific feature dimensions, directly linking signal intensity to stylistic emphasis and motion amplitude. 

Second, to allow the transformer's self-attention mechanism to explicitly resolve high-frequency temporal dependencies between the multimodal condition signals and the motion cues, we concatenate the local condition features $\mathbf{c}$ with the modulated motion $\mathbf{h}'_t$ along the feature dimension. The resulting augmented feature representation, $\mathbf{\tilde{h}}_t$, passed to the subsequent self-attention layers, is formally defined as:
  $\mathbf{\tilde{h}}_t = \bigl[ \mathbf{h}'_t \,;\, \mathbf{c} \bigr].$

\textbf{Long-form Generation.}
\label{par:long-form}
To synthesize motion sequences of arbitrary lengths, we partition the generation process into overlapping windows of length $W$ with an overlap of $H$ frames between consecutive windows. Following the Double-Take strategy~\cite{shafir2024human}, the diffusion model processes all windows concurrently within a single shared batch.  To enforce temporal continuity and prevent sequence drift, we perform a blending operation (a ``handshake'') across the overlapping regions at the conclusion of each denoising step $t$. 
Let $\mathbf{x}^{(k)}_t$ denote the intermediate noisy motion sequence for the $k$-th window at diffusion step $t$. For the overlapping region of length $H$, we linearly interpolate between the tail of window $k-1$ and the head of window $k$ using weights $w_i = i/H$:
\[
\tilde{\mathbf{x}}_t[i] = (1-w_i)\,\mathbf{x}^{(k-1)}_t[W-H+i] + w_i\,\mathbf{x}^{(k)}_t[i].
\]
We overwrite the boundary regions in both adjoining windows with the blended result $\tilde{\mathbf{x}}_t$, effectively preventing temporal drift and discontinuities.

\subsection{Part II: Joint Motion Integration (JMI)}

We propose Joint Motion Integration (JMI), a training-free mechanism that injects retrieved long-tail semantic gestures directly into the co-speech diffusion sampling process. By utilizing the spatial variance of the source motion, JMI intervenes in the predicted clean motion $\hat{\mathbf{x}}_0$ at each denoising step. It enforces kinematic constraints on specific joints while allowing the personalized diffusion backbone prior to synthesize plausibly stylized and natural motion for the remaining body parts.

\textbf{Active-Joint Masking.}
Each source semantic gesture is represented by a motion sequence in explicit rotation space, $G_m \in \mathbb{R}^{6J \times N_m}$, where $N_m$ denotes the sequence length in frames and $J$ is the number of joints. A naive injection of a full-body \ar{(all joints)} source motion clip (i.e., naive in-betweening) would inevitably overwrite the target speaker's unique identity. To prevent this, we operate on the observation that a gesture's semantic meaning is typically conveyed only by pronounced movements in a localized subset of joints. By \ar{automatically} isolating and injecting \textit{only} these specific joints, we can preserve the target speaker's overarching style. To automate this, we pre-compute an \emph{active-joint mask}, $M_m \in \{0,1\}^{6J}$, for each gesture by thresholding its kinematic variance. Joints exhibiting significant motion, along with their kinematic parents, are marked as active, while the remaining joints (low-variance) are masked as non-active. \ar{Manual masks are supported as an optional fallback for edge cases such as near-static source poses.}

\ar{\textbf{Semantic Gesture Selection.}
To determine which semantic gesture to inject and when to inject it, we first align the speech signal with the transcript using a speech-to-text model that provides word-level timestamps. This yields a sequence $\{(w_i, s_i, e_i)\}_{i=1}^{M},$
where $w_i$ is the $i$-th word and $s_i, e_i$ denote its start and end times. Given the transcript, candidate semantic gestures may be selected either manually or automatically. In our current pipeline, this stage is performed automatically using a zero-shot LLM call. The LLM receives the transcript together with our semantic gesture collection  (each gesture defined by a name, a textual description, and an example), and returns an augmented transcript with inline gesture annotations, e.g., ``The police shot \texttt{(HAND\_SHOOT)}'', associating the gesture \texttt{HAND\_SHOOT} with the word ``shot''. \ar{The exact zero-shot LLM prompt and additional transcript tagging examples are provided in the supplementary material.}

For each annotated word $w_k$, we retrieve the corresponding gesture clip $G^m$ of length $L_m$ and its precomputed active-joint mask $M^m$. We then center it at the midpoint of the annotated word: \[
[s_k^*, e_k^*]
=
\left[
\frac{s_k + e_k}{2} - \frac{L_m}{2},
\;
\frac{s_k + e_k}{2} + \frac{L_m}{2}
\right].
\]

This produces a set of injection specifications
\[
\mathcal{I}
=
\bigl\{\,(M^m, s_k^*, e_k^*, G^m, \lambda_{\mathrm{inj}}, B_{\mathrm{blend}})\,\bigr\},
\]
where $\lambda_{\mathrm{inj}} \in [0,1]$ controls the injection strength and $B_{\mathrm{blend}}$ determines the number of frames used for temporal blending near the interval boundaries. }

\textbf{Per-step Joint-Space Intervention.}
During inference, at each diffusion step $t$, the backbone denoiser predicts the clean motion $\hat{\mathbf{x}}_0 = f_\theta(\mathbf{x}_t, t, \mathrm{cond})$ for every temporal window. For each tagged gesture intersecting the current window (indexed by $w$), we project and crop it to the window to obtain $G^m_{\mathrm{crop}}$ of length $L$, then blend the gesture into the prediction using a mixing matrix \[\bm{\Phi} = \lambda_{\mathrm{inj}} \left(M^m \otimes \tau\right) \in [0,1]^{6J \times L},\] where $M^m$ restricts the edit to the active joints and $\tau$ is a trapezoidal profile that ramps smoothly
at the boundaries:
    \[
    \hat{\mathbf{x}}_0^{\text{edit}}[:6J, s_k^w:e_k^w] \leftarrow
    (\mathbf{1} - \bm{\Phi}) \odot \hat{\mathbf{x}}_0[:6J, s_k^w:e_k^w]
    + \bm{\Phi} \odot G^m_{\mathrm{crop}}.
    \]
\ar{Intuitively, $\Phi$ is nonzero only on the active joints and frames of the gesture, so those joints are pulled toward the target while all others keep the diffusion model's prediction, preserving the speaker's style and natural motion.
The modified prediction $\hat{\mathbf{x}}_0^{\text{edit}}$ replaces $\hat{\mathbf{x}}_0$ in the standard DDIM posterior sampling step to compute the subsequent state $\mathbf{x}_{t-1}$. (Exact window projection, cropping, and the construction of $\tau$ are detailed
in the Supplementary Material.)}

\textbf{Generalization and Composition.}
JMI is source-agnostic; any motion sequence in the SMPL-X format can be injected, if accompanied by appropriate timing annotations. Furthermore, the joint-wise masking supports compositional editing, allowing multiple non-conflicting gestures (e.g., a lower-body jump concurrent with an upper-body wave) to be injected simultaneously.

\subsection{Implementation Details}
\label{sec:method-impl}

We represent motion using a 55-joint SMPL-X rig sampled at 30 FPS, processed in temporal windows of size $W{=}196$. Our backbone is an 8-layer Transformer with a hidden dimension of 512, pre-trained on the AMASS dataset~\cite{mahmood2019amass} and subsequently fine-tuned on BEAT2~\cite{liu2024emage}. The network is optimized using AdamW with a learning rate of $10^{-4}$. We set the loss weights to $\lambda_{\text{vel}}{=}100$ and $\lambda_{\text{fc}}{=}50$. Training was conducted on a single NVIDIA RTX 3090 GPU. The initial base training converged in approximately 24 hours, while identity-specific fine-tuning required roughly 90 minutes per subject. During inference, we employ DDIM sampling with $T{=}1000$ timesteps and a handshake overlap of \ar{$H{=}30$} frames. For our JMI framework, we apply a default injection strength of $\lambda_{\text{inj}}{=}0.65$ alongside a blend window of $B_{\text{blend}}{=}10$. To ensure a fair evaluation, all co-speech gesture generation baselines were evaluated on an identical test set, and our JMI framework is compared against injection baselines using the exact same gesture annotations.
\section{Datasets \& Preprocessing}
\label{sec:datasets}

\textbf{AMASS (Pretraining).} 
We utilize AMASS~\cite{mahmood2019amass} to learn a robust prior for human motion. This archive unifies over 40 hours of motion capture data from more than 300 subjects and $11{,}000$ sequences. Although AMASS does not contain speech, its scale allows our diffusion backbone to learn realistic full-body kinematics and contact dynamics. We use this diverse data to pretrain a generalized motion prior before fine-tuning on specific speaker identities.

\textbf{BEAT2 (Fine-tuning).} 
For co-speech gesture generation, we employ the BEAT2 dataset~\cite{liu2022beat,liu2024emage}, a large-scale corpus capturing 30 speakers in conversational scenarios. To achieve high-fidelity personalization, we fine-tune our AMASS-pretrained backbone separately for each target identity. For our experiments, we selected five distinct speakers covering a range of styles: \emph{Scott}, \emph{Ayana}, \emph{Wayne}, \emph{Lawrence}, and \emph{Sophie}. We utilize approximately 90 minutes of training data per subject.

\textbf{Enhanced Transcript Alignment.} 
To ensure precise semantic conditioning, we improved the quality of the textual annotations in BEAT2. We re-transcribed the selected subsets using the Whisper Automatic Speech Recognition (ASR) model, yielding significantly more accurate transcripts than the original release. 

\textbf{SeG (Semantic Injection).} 
As a proof of concept for our Joint Motion Integration (JMI) module, we construct an SMPL-X semantic gesture collection, extracted from the SeG dataset~\cite{zhang2024semantic}, a curated subset of Semantic Gesticulator containing over 200 distinct semantic gestures (including iconic, metaphoric, and deictic motions). However, we emphasize that our pipeline is fundamentally source-agnostic and can seamlessly integrate any arbitrary ``in-the-wild'' motion sequence.

\textbf{BVH$\rightarrow$SMPL-X Retargeting.} 
We retarget all SeG BVH clips to the SMPL-X parameterization using a custom pipeline that includes joint mapping, scale calibration, and Inverse Kinematics (IK) based cleanup. Note that this step is required only because the source data is in BVH format; if the semantic motions were originally in SMPL-X format, this retargeting would not be needed. More details about the retargeting process are in the Supplementary Material.

\section{Experiments}
\label{sec:experiments}

We validate SiGnature (our co-speech backbone and JMI framework) against state-of-the-art baselines using both quantitative and perceptual metrics. Since static figures cannot fully capture temporal coherence, we strongly encourage readers to view the supplementary video to observe our results in motion.

\subsection{Metrics}
\label{sec:metrics }

We evaluate motion quality by focusing on three primary categories: standard generative metrics, kinematic quality, and semantic fidelity.

\textbf{1. Standard Generative Metrics.}
We employ community-standard metrics to evaluate the distribution, variation, and synchrony of co-speech motion generation.
\textbf{Fréchet Gesture Distance (FGD)}~\cite{yoon2020trimodal} measures the feature-space distance between generated and ground-truth gesture distributions.
\textbf{Diversity}~\cite{li2021audio2gestures} computes the variability of the generated motions.
\textbf{Beat Consistency (BC)}~\cite{siyao2022bailando} measures the synchronization between audio beats and kinematic motion beats.

\textbf{2. Kinematic Smoothness.}
Standard benchmarks often ignore the physical quality of motion, yet high-frequency jitter is a common artifact in diffusion models.
Minimizing jerk is considered the gold standard for natural movement in robotics~\cite{flash1985coordination,berret2011evidence}.
We quantify this using \textbf{Mean Absolute Jerk} and \textbf{Mean Absolute Acceleration} in local joint space.

\textbf{3. Semantic Fidelity and Style Preservation.}
We assess our Joint Motion Integration (JMI) approach using a comprehensive set of metrics covering reconstruction quality, semantic fidelity, and identity preservation. \textbf{Reconstruction Fidelity (Baselines)} measures the \textbf{Mean Joint Error} (cm) between a source semantic gesture and its reconstruction via baseline VQ-VAE codebooks (Table~\ref{tab:vqvae_reconstruction}). 
\textbf{Semantic Accuracy} is determined via a human perceptual study (Fig.~\ref{fig:user_study_semantic}).
\textbf{Injection Fidelity} reports Mean Joint Error (cm) regarding both \textit{active joints} (accuracy) and \textit{non-active joints} (style) (defined in the Supplementary Material). 
Furthermore, we measure \textbf{Style Preservation} using Cross-Identity FGD between personalized outputs and target ground-truth (Table~\ref{tab:jmi_personalization}).

\subsection{Baselines}
\label{sec:experiments-baselines}
We evaluate our framework in two capacities: our backbone as a high-fidelity co-speech gesture generator, and JMI as a semantic injection system.

\textbf{Co-Speech Generation Quality.}
To validate the quality of our Base Diffusion Backbone (prior to injection), we compare against state-of-the-art co-speech gesture models: \textbf{EMAGE}~\cite{liu2024emage}, \textbf{SynTalker}~\cite{chen2024syntalker}, and \textbf{GestureLSM}~\cite{liu2025gesturelsm}.

\textbf{Semantic Injection and Preservation.}
To evaluate semantic fidelity, we compare our \textbf{Full Model (with JMI)} against \textbf{Semantic Gesticulator}~\cite{zhang2024semantic}. Additionally, we introduce two specific baselines to validate the necessity of explicit-space semantic injection and active masking:
\begin{itemize}
\ar{\item \textbf{Naive In-betweening (No Masking):} A baseline that injects the source gesture into \textit{all} body joints during the diffusion process. Because the underlying injection mechanism is identical to our full method, this comparison isolates the impact of the active-joint mask, demonstrating its necessity for preserving the target speaker's personal style.}
\item \textbf{SynTalker VQ-VAE (Reconstruction):} To quantify the information loss inherent in latent spaces, we evaluate the reconstruction fidelity of semantic gestures passed through the VQ-VAE codebooks of SynTalker~\cite{chen2024syntalker}. This validates our hypothesis that discrete latent representations erode fine-grained semantic details.
\end{itemize}

\begin{table}[t]
  \centering
\caption{\textbf{Quantitative comparison and ablations.} 
We report standard metrics: FGD ($\times 10^{-1}$), Diversity and BC ($\times 10^{-1}$), which capture distributional realism, semantic diversity, and audio--motion alignment. 
We additionally report kinematic smoothness (Jerk/Accel) and generation speed as frames per second (FPS). 
\textit{gLSM-D} and \textit{gLSM-MF} denote the diffusion and MeanFlow variants of GestureLSM. 
Rows with \textit{(steps)} indicate varying DDIM sampling iterations to test efficiency trade-offs.}
  \label{tab:quantitative-main}
  \resizebox{\linewidth}{!}{%
  \begin{tabular}{lcccccc}
    \toprule    
    \multirow{2}{*}{\textbf{Method}} &
    \multicolumn{3}{c}{\textbf{Standard Metrics}} &
    \multicolumn{2}{c}{\textbf{Smoothness (XYZ)}} &
    \multirow{2}{*}{\textbf{FPS}$\uparrow$} \\
    & \textbf{FGD}$\downarrow$ & \textbf{Diversity}$\uparrow$ & \textbf{BC}$\uparrow$ &
      \textbf{Jerk}$\downarrow$ & \textbf{Accel}$\downarrow$ & \\
    \midrule
    EMAGE                    &  5.54  & 13.08  & \textbf{7.70}  &  66.76   &  4.16   &  925 \\
    SynTalker                &  4.66  & 12.30  & 7.37  &  55.40   &  3.40   &   18 \\
    gLSM-D   &  4.10  & 12.56  & 7.35  &  57.73   &  3.47   &  399 \\
    gLSM-MF    &  \textbf{4.04}  & 12.46  & 7.45  &  61.16   &  3.69   & 1890 \\
    GT                       &  0.00  & 13.09  & 6.83  &  42.13 & 3.03 &   -- \\
    \midrule
    Ours (1k steps) & 4.72 & \textbf{13.66} & 6.81
                               & 34.91 & 2.60 & 55 \\
    Ours (100 steps)         &  5.49  & 13.52  & 6.81  &  \textbf{33.92} & \textbf{2.57} & 547 \\
    Ours (20 steps)          & 5.48  & 13.29  & 6.84  &  33.99 &  2.59 & \textbf{2802} \\
    Ours w/o CLIP\&FiLM    &  7.28  & 13.08  & 6.86  &  34.93 &  2.67 &  -- \\
    Ours w/o AMASS           &  7.34  & 10.07  & 7.46  &  69.65 & 3.88 &  -- \\
    \bottomrule
  \end{tabular}
  }
\end{table}


\begin{table}[t]
  \centering
\caption{\textbf{Impact of Active-Joint Masking and Per-Step Intervention.} We compare our selective, per-step injection strategy (Ours w/ JMI) against two baselines: (1) a naive in-betweening approach that forces the source semantic motion onto \textit{all} body joints across all diffusion steps (Ours w/ IB), and (2) a variant that uses active-joint masking but only applies a single-step intervention at the final step T=0 (Ours w/ JMI Post Integration). The high FGD for the naive baseline indicates that injecting motion without masking deviates significantly from the speaker's ground-truth style. Furthermore, the single-step intervention degrades kinematic smoothness compared to continuous harmonization. Our full JMI effectively recovers the FGD and enhances diversity by preserving the target's style in non-active joints.}
  \label{tab:quantitative-sem}
  \resizebox{\linewidth}{!}{%
  \begin{tabular}{lcccccc}
    \toprule    
    \multirow{2}{*}{\textbf{Method}} &
    \multicolumn{3}{c}{\textbf{Standard metrics}} &
    \multicolumn{2}{c}{\textbf{Smoothness (XYZ)}} &
     \\
    & \textbf{FGD}$\downarrow$ & \textbf{Diversity}$\uparrow$ & \textbf{BC}$\uparrow$ &
      \textbf{Jerk}$\downarrow$ & \textbf{Accel}$\downarrow$ & \\
    \midrule
    Ours w/ IB                & 22.54  & 14.58  & 7.49  &  43.23 & 3.24 \\
    Ours w/ JMI Post Integration                & 16.28  & 14.66  & 7.76  &  44.83 & 3.34 \\
    Ours w/ JMI               & \textbf{11.88}  & \textbf{14.10}  & \textbf{7.74}  &  \textbf{40.31} &  \textbf{3.03} \\
    \bottomrule
  \end{tabular}
  }
\end{table}

\begin{table}[t]
\centering

\label{tab:vqvae_reconstruction}
\caption{\textbf{Latent Reconstruction Fidelity.} We evaluate the ability of the SOTA VQ-VAE baseline \cite{chen2024syntalker} to reconstruct motion. While it accurately reconstructs in-distribution conversational motion (BEAT2), the error quadruples on out-of-distribution semantic gestures (SeG), particularly on the active joints carrying the semantic meaning.}
\label{tab:vqvae_reconstruction}
\resizebox{\columnwidth}{!}{
\begin{tabular}{llc}
\toprule
\textbf{Evaluation Set} & \textbf{Evaluated Region} & \textbf{MPJPE (cm)} $\downarrow$ \\
\midrule
BEAT2 & Full Body & $5.34 \pm 0.85$ \\
\midrule
SeG & Full Body & $16.32 \pm 3.37$ \\
SeG & \textbf{Active Joints} & $\mathbf{21.64 \pm 5.64}$ \\
\bottomrule
\end{tabular}
}
\end{table}

\subsection{Quantitative Results}
\label{sec:quantitative}
We present our comparative results in Table~\ref{tab:quantitative-main} and Table~\ref{tab:quantitative-sem}. Our analysis reveals that \emph{SiGnature} achieves a superior balance between kinematic quality and expressive range.

\begin{figure}[!t]
  \centering
  \includegraphics[width=0.49\textwidth]{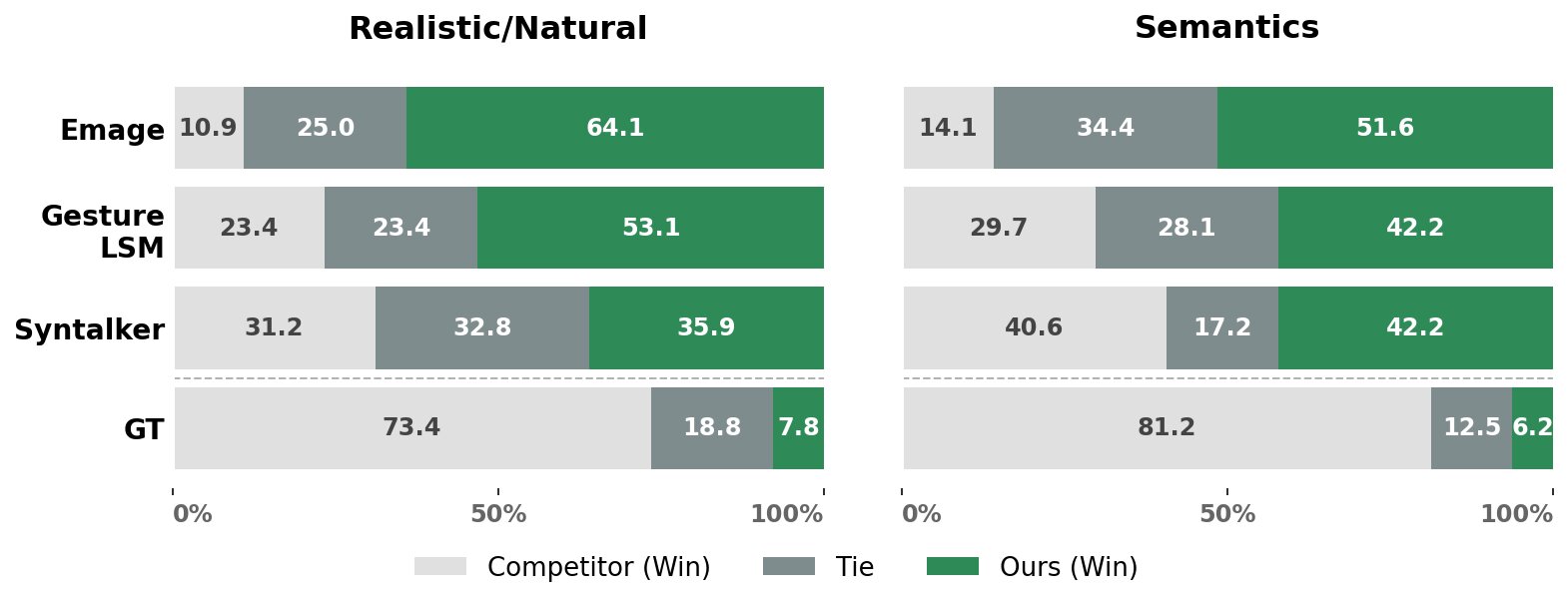}
\caption{\textbf{Perceptual Study Results (N=64).}
Values indicate the percentage of user preferences.
Green bars denote a preference for our method. Users consistently favored the smoothness of our rotation-space approach over the latent-space baselines.}
\label{fig:userstudy-backbone}
\end{figure}

\textbf{1. Co-Speech Performance (Backbone Analysis).}
\label{par:human-performance}
While latent-based co-speech baselines like GestureLSM achieve a marginally better FGD score (4.10 vs.\ our 4.72), this slight numerical gap highlights a known lack of correlation between feature-space metrics and human perception~\cite{kucherenko2024evaluating}. In practice, latent models achieve slightly better FGDs but suffer from severe kinematic instability and reconstruction errors. Quantitatively, latent baselines like SynTalker, GestureLSM, and EMAGE exhibit high-frequency jitter, reflected in exceptionally high Jerk scores (55.40, 57.73, and 66.76, respectively), significantly exceeding the Ground Truth (42.13). By generating gestures directly in explicit space, our architecture \ar{substantially reduces} this jitter, achieving the lowest Jerk (\textbf{34.91}) and Acceleration (\textbf{2.60}). Our perceptual studies (Fig.~\ref{fig:userstudy-backbone}) confirm that human raters prefer SiGnature's physical naturalness over baselines, despite the small FGD gap. Furthermore, our \textbf{Diversity} score (\textbf{13.66}) exceeds the Ground Truth (13.09). This combination (low jerk and high diversity), demonstrates that explicit rotation-space diffusion effectively removes high-frequency noise while retaining a large-scale expressive range. We term this \emph{Active Smoothness}: motion that is physically clean and fluid, yet highly dynamic.

\textbf{2. Semantic Injection Performance.}
We evaluate the effectiveness of our Joint Motion Integration (JMI) against naive approaches and latent baselines.

\begin{figure*}[!t]
\centering
\scalebox{1}{%
\begin{minipage}{\textwidth}
\scriptsize
\begin{minipage}[t]{0.04\textwidth} 
\raggedleft
\vspace*{0.10\textheight}
\raisebox{0.5\height}{\rotatebox{90}{\textbf{Lawrence}}}\par\vspace*{0.08\textheight}
\vspace*{0.10\textheight}
\raisebox{0.5\height}{\rotatebox{90}{\textbf{Ayana}}}\par\vspace*{0.08\textheight}
\vspace*{0.10\textheight}
\raisebox{0.5\height}{\rotatebox{90}{\textbf{Wayne}}}\par\vspace*{0.08\textheight}
\vspace*{0.10\textheight}
\raisebox{0.5\height}{\rotatebox{90}{\textbf{Scott}}}
\end{minipage}
\hfill
\begin{minipage}[t]{0.46\textwidth} 
\centering
\textbf{Beast Claw Gesture}\par
\vspace{1mm}
\makebox[0.33\linewidth][c]{\textbf{Source motion}}%
\makebox[0.33\linewidth][c]{\textbf{JMI-injected result}}%
\makebox[0.33\linewidth][c]{\textbf{Base motion}}
\vspace{1mm}
\includegraphics[width=\linewidth,keepaspectratio]{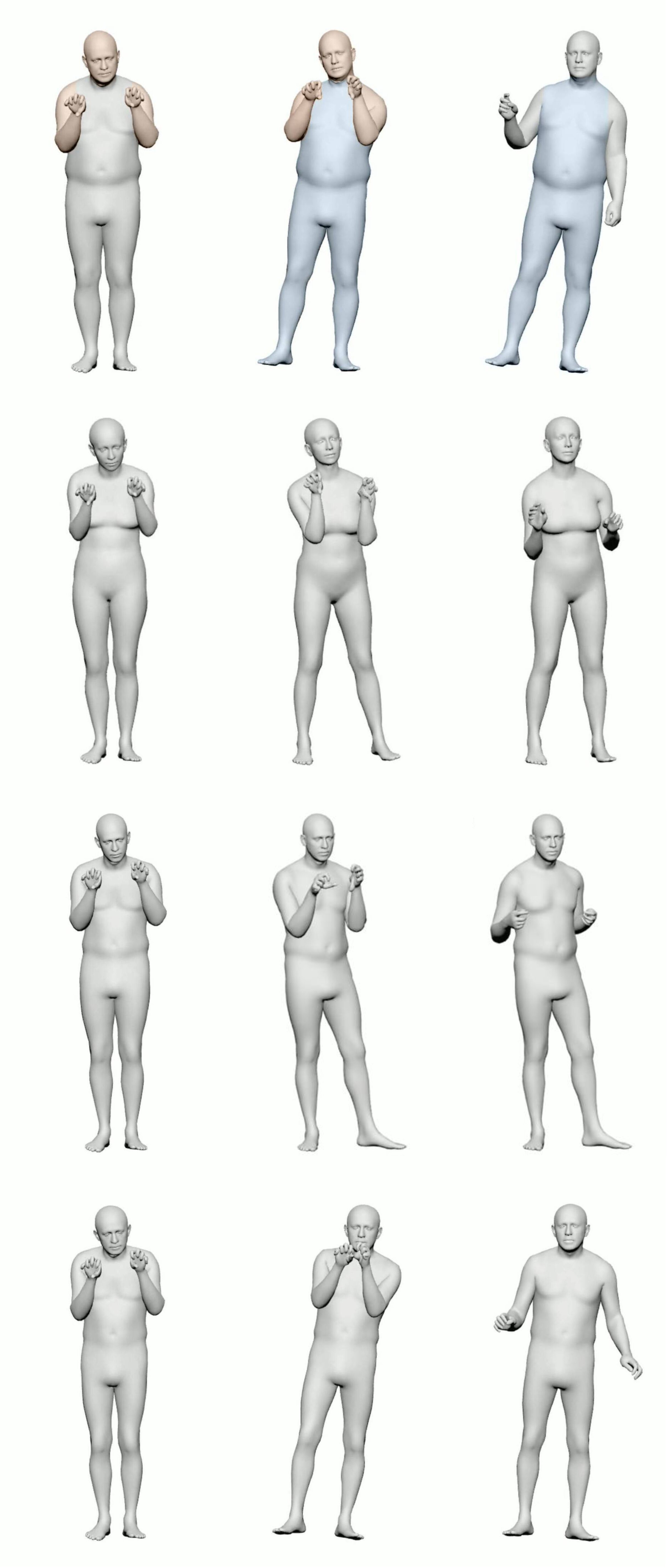}
\end{minipage}
\vrule
\begin{minipage}[t]{0.46\textwidth} 
\centering
\textbf{Checking the Time Gesture}\par
\vspace{1mm}
\makebox[0.33\linewidth][c]{\textbf{Source motion}}%
\makebox[0.33\linewidth][c]{\textbf{JMI-injected result motion}}%
\makebox[0.33\linewidth][c]{\textbf{Base motion}}
\vspace{1mm}
\includegraphics[width=\linewidth,keepaspectratio]{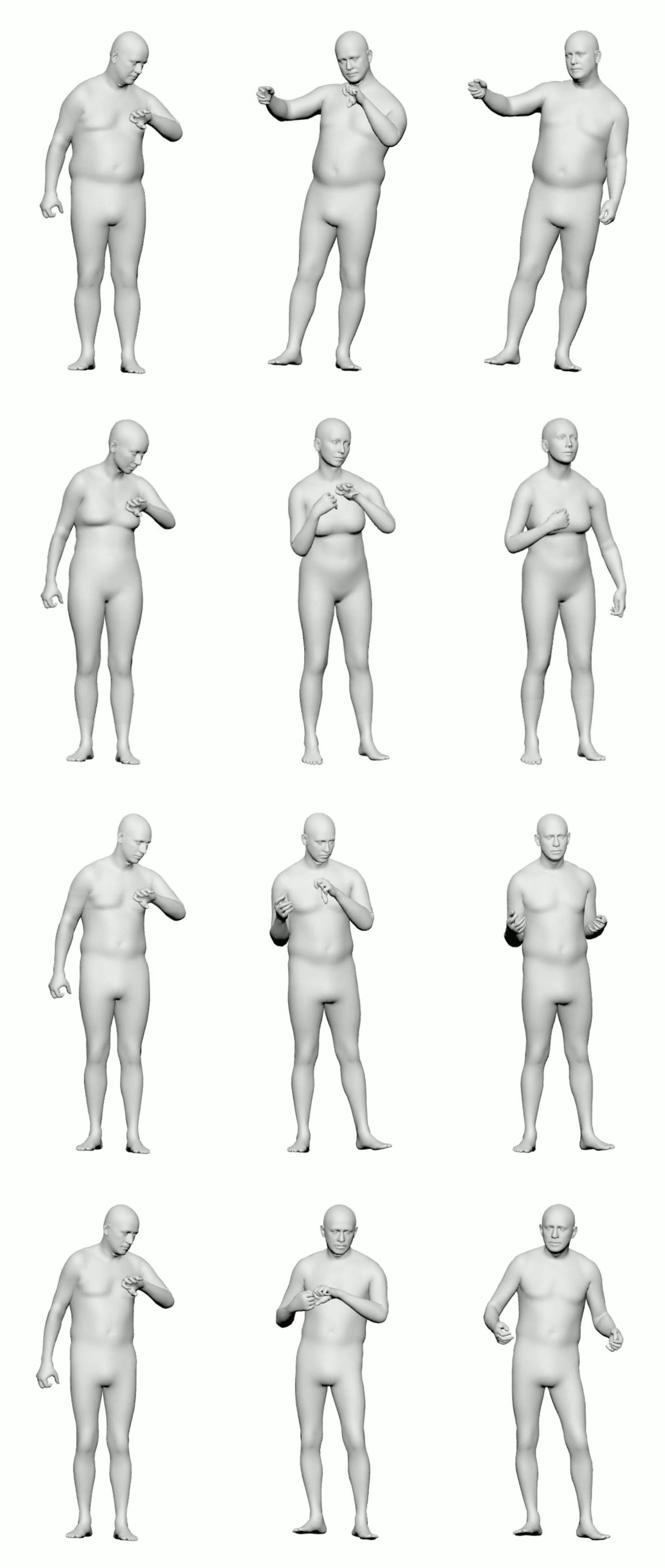}
\end{minipage}
\end{minipage}%
}
\caption{\textbf{Cross-identity semantic gesture injection.}
We show two semantic gestures, \textbf{Beast Claw} (left) and \textbf{Checking the Time} (right) applied to multiple target speakers.
For each gesture, we compare the \textbf{source semantic motion} (gesture bank), our \textbf{JMI-injected result}, and the \textbf{base diffusion output} (no injection).
JMI modifies only the active joints, preserving each speaker's posture and motion style while enforcing the intended semantics.}
\label{fig:qualitative1}
\end{figure*}

\textbf{Benefit of JMI vs. Naive In-Betweening.}
Table~\ref{tab:quantitative-sem} demonstrates the necessity of our masking strategy. The ``Naive In-betweening'' baseline, which forces source motion onto all joints, results in a severe degradation of distribution realism (FGD rises to 22.54). Figure~\ref{fig:qualitative1} (top-left) visualizes this advantage: while the source motion features an upright stance, JMI successfully injects the ``Beast Claw'' hand gesture into the target speaker (Lawrence) while preserving his characteristic leaning posture in the lower body. By selectively injecting only active joints, JMI recovers the natural distribution and increases semantic diversity.

\ar{
\textbf{Per-Step vs. Single-Step Intervention.} A natural question is whether the gesture can be injected once into the final clean prediction rather than at every denoising step. We
tested a single-step variant applying JMI only at T=0 (posthoc) using the same blending, removing per-step harmonization. Without continuous intervention, the diffusion prior cannot progressively adapt the non-active joints to physically support the injected gesture, producing visible kinematic artifacts at the injection boundaries. Quantitatively, the per-step intervention (FGD 11.88, Jerk 40.31) substantially outperforms the single-step injection (FGD 16.28, Jerk 44.83), confirming that per-step harmonization is essential.
}

\textbf{Superiority of Explicit Space.}
Table~\ref{tab:vqvae_reconstruction} highlights the limitations of latent baselines. While the SynTalker VQ-VAE accurately reconstructs in-distribution conversational motion (Error: 5.34cm), the error quadruples on out-of-distribution semantic gestures (16.32cm), particularly on the active joints carrying the semantics (21.64cm). \ar{For direct comparison, JMI attains a faithful active-joint-to-target alignment of just 12.9 cm for Scott (Supplementary, Table 5) on the same gestures}. This confirms that discrete latent spaces struggle to preserve the fine-grained spatial details required for long-tailed semantic gestures, justifying our choice of explicit rotation-space diffusion.

\textbf{Style Preservation and Identity Leakage.} 
\ar{Style is captured implicitly by fine-tuning a separate personalized diffusion model per identity (Sec. 4).} Table~\ref{tab:jmi_personalization} evaluates our framework's ability to preserve target speaker identity during semantic injection, \ar{preventing overwriting the speaker’s distinctive signature style.} The consistently low diagonal entries confirm that SiGnature successfully synthesizes the target speaker’s gestural style. The off-diagonal entries remain comparatively high, demonstrating that our model does not suffer from style leakage or mode collapse into an "average" speaker. Furthermore, the last row (\textit{Source SeG}) reveals that the raw semantic gestures originate from a distinct, out-of-distribution style. By contrasting these high source FGDs with our low diagonal scores (e.g., Sophie: 5.19 $\to$ 1.87), we quantitatively demonstrate that JMI avoids naive motion-pasting; instead, it seamlessly integrates the semantic action while pulling the integrated semantic gesture back into the target speaker's specific stylistic manifold.

\textbf{Inference Speed.} As detailed in Table~\ref{tab:quantitative-main}, our non-autoregressive architecture offers a throughput advantage. By leveraging the Double-Take strategy (Sec.~\ref{par:long-form})~\cite{shafir2024human} to batch full utterances, we achieve \textbf{55 FPS} at 1,000 steps. Using 20 steps, we reach \textbf{2,802 FPS} while maintaining high performance. This represents a massive throughput advantage over baselines like SynTalker (18 FPS). See Sec.~\ref{sec:method-impl} for technical details. Furthermore, because JMI introduces negligible computational overhead, it preserves our backbone's real-time generation capabilities.

\vspace{-3mm}

\subsection{Ablations}
\label{sec:experiments-ablations}
To validate our backbone architectural choices, we evaluate three variants of our model (Table~\ref{tab:quantitative-main}).

\textbf{Conditioning Mechanisms.}
Removing the CLIP and FiLM conditioning (``Ours w/o CLIP\&FiLM'') leads to a significant degradation in FGD ($4.72 \to 7.28$), while other metrics remain largely unaffected. This confirms that our dual-stream architecture (global sentence semantics + local token prosody) is specifically responsible for aligning the motion distribution with the target style, without compromising kinematic quality. \ar{Additional ablations isolating the effect of the global CLIP prior, including CLIP-only conditioning and inference-time prompt manipulation, are detailed in the supplementary material.}

\textbf{Motion Prior (AMASS).}
Training directly on BEAT2 without AMASS pretraining (``Ours w/o AMASS'') results in a drastic spike in Jerk ($34.91 \to 69.65$). This validates the importance of learning a robust, large-scale motion prior for ensuring kinematic smoothness and stability.






\begin{figure}[t]
  \centering
  \includegraphics[width=0.42\textwidth]{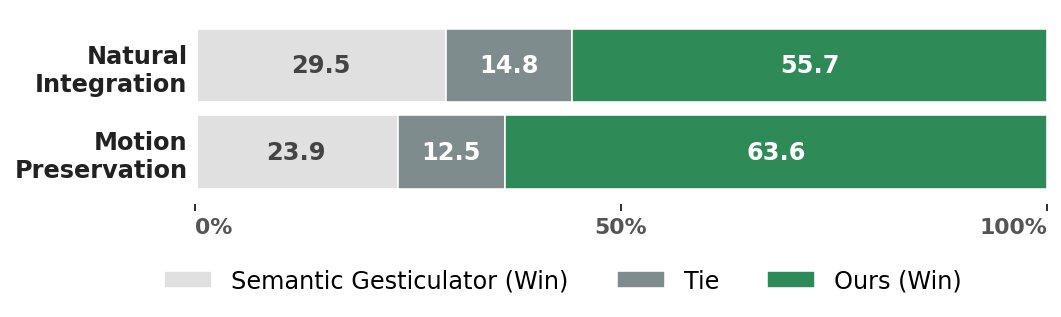}
\caption{\textbf{Perceptual Study II: Semantic Gesture Integration.} 
We compare our method against \textit{Semantic Gesticulator}~\cite{zhang2024semantic} ($N=88$). Values indicate the percentage of user preferences. 
Green bars denote our method. 
}

\label{fig:user_study_semantic}
\end{figure}

\begin{table}[t]
\centering
  \caption{\textbf{Cross-identity FGD for JMI personalization.}
  Columns denote the ground-truth (GT) motions, rows are our results.
  Lower is better. Diagonal entries measure same-identity; off-diagonals measure style leakage across identities.
  The last row reports our semantic motion collection. (Scores are $\times 10$.)}
  \label{tab:jmi_personalization}
\setlength{\tabcolsep}{3.5pt} 
\renewcommand{\arraystretch}{1.3} 
\resizebox{\columnwidth}{!}{
\begin{tabular}{lccccc}
\toprule
\textbf & \textbf{Scott (GT)} & \textbf{Wayne (GT)} & \textbf{Ayana (GT)} & \textbf{Sophie (GT)} & \textbf{Lawr. (GT)} \\
\midrule
Scott (Gen.) & \textbf{1.00} & 3.07 & 3.66 & 3.55 & 2.92 \\
Wayne (Gen.) & 1.81 & \textbf{1.28} & 3.35 & 2.84 & 2.77 \\
Ayana (Gen.) & 2.72 & 3.06 & \textbf{1.84} & 3.40 & 3.41 \\
Sophie (Gen.) & 3.04 & 3.88 & 4.09 & \textbf{1.87} & 4.24 \\
Lawr. (Gen.) & 1.36 & 2.41 & 3.11 & 3.07 & \textbf{1.18} \\
\midrule
\textit{Source (SeG)} & \textit{3.68} & \textit{4.36} & \textit{4.57} & \textit{5.19} & \textit{3.22} \\
\bottomrule
\end{tabular}
}
\end{table}

\subsection{Perceptual Study}
\label{sec:experiments-perceptual}
We conducted three studies to evaluate co-speech gesture generation quality, semantic integration, and explicit motion benefits. Statistical significance was determined using one-sided binomial tests (for additional details see the Supplementary Material).

\textbf{Study I: Co-Speech Naturalness (Only Our Co-Speech Generator Backbone).}
We recruited 32 participants to conduct a side-by-side preference test ($N=64$ pairwise comparisons) assessing realism and semantic correctness, comparing our method to EMAGE, GestureLSM, SynTalker, and ground truth. Overall, our method received higher ratings than all three baselines across both criteria. As shown in Figure~\ref{fig:userstudy-backbone}, these differences were statistically significant relative to EMAGE on both metrics ($p < 0.001$) and relative to GestureLSM in realism ($p < 0.01$), while the remaining differences, including comparisons with SynTalker, were not statistically significant. These results suggest that kinematic smoothness contributes meaningfully to perceived human-likeness (and not necessarily the FGD, see Sec.~\ref{par:human-performance}).

\textbf{Study II: Semantic Injection Quality (Backbone + JMI).}
We compared our Full Model against \textit{Semantic Gesticulator} ($N=88$ evaluations from 22 participants) using skeleton  visualizations, utilizing the exact same gesture annotations. Results (Figure~\ref{fig:user_study_semantic}) show our method was rated significantly higher in \textbf{Motion Preservation} ($p < 0.001$) and \textbf{Natural Integration} ($p < 0.025$). Viewers found that our explicit injection preserved fine-grained details better while maintaining high naturalness.

\textbf{Study III: Perceived Impact of JMI (Full Model vs. Backbone).}
This study validated JMI's critical impact on perceived quality by comparing our Full Model (with JMI-injected gestures) against the Diffusion Backbone alone. Participants ($N=64$) showed a strong preference for the Full Model, favoring it for \textbf{Naturalness} (63\% vs.\ 37\%) and significantly for \textbf{Semantic Accuracy} ($61\%$ vs.\ $39\%$, $p < 0.05$). This confirms that JMI's explicit semantic gestures effectively break the monotony of standard beat gestures, leading to more lively, engaging, and realistic co-speech generation.
\section{Conclusions}

\label{sec:conclusion}
We have presented SiGnature, a framework for co-speech gesture generation that reconciles precise semantic control with high-fidelity style preservation. Extensive quantitative and perceptual studies demonstrate that our explicit-space diffusion backbone achieves state-of-the-art kinematic smoothness, producing more natural and human-like motion than existing latent-based methods. Furthermore, our Joint Motion Integration (JMI) mechanism effectively decouples semantic intent from speaker identity, allowing users to seamlessly inject rare, in-the-wild semantic gestures directly into the generated sequence.

Despite these strengths, we acknowledge specific limitations in our current implementation. First, our approach focuses exclusively on body and hand dynamics. While it does not natively address facial expression generation, it could be readily integrated with existing state-of-the-art facial synthesis models. Second, because our model processes motion in overlapping temporal windows, it is not strictly real-time. Latency is determined by the window size and the number of diffusion steps, requiring a trade-off between inference speed and final motion quality. Finally, identifying ``active joints'' based on a spatial variance heuristic is effective but simplistic; it may occasionally miss subtle, low-amplitude semantic movements or require manual adjustment when extracting gestures from highly complex or noisy source clips. \ar{We also note that word-level timestamps are a coarse approximation and may not capture sub-word audio beats or fine intra-word timing, which can slightly offset the centering of very short gestures.}

Future work can address these constraints by replacing the variance-based heuristic with a learned, semantic-aware predictive network for joint selection. Furthermore, integrating our JMI approach into broader speech-to-gesture pipelines could fully automate the generation of long-tailed semantic motions. This could involve coupling our backbone with a semantic gesture retrieval system or a dedicated text-to-semantic-gesture module, ultimately helping to overcome the scarcity of semantic gestures in existing motion capture datasets. We also aim to extend our explicit-space co-speech diffusion architecture into a training-free multi-speaker, or environment-aware model \ar{and achieve finer-grained alignment}.

In summary, our work highlights that while highly compressed latent spaces excel at learning broad statistical distributions, they fundamentally struggle with rare, fine-grained, and long-tailed semantic gestures. By operating directly in explicit rotation space and leveraging targeted, training-free interventions, we can overcome dataset limitations and achieve a level of physical realism, editability, and localized control that purely end-to-end latent models cannot yet match.
\begin{figure*}[t]
\centering
\scriptsize

\begin{minipage}[t]{0.04\textwidth} 
\raggedleft
\vspace*{0.10\textheight}
\raisebox{0.5\height}{\rotatebox{90}{\textbf{Lawrence}}}\par\vspace*{0.08\textheight}
\vspace*{0.10\textheight}
\raisebox{0.5\height}{\rotatebox{90}{\textbf{Ayana}}}\par\vspace*{0.08\textheight}
\vspace*{0.11\textheight}
\raisebox{0.5\height}{\rotatebox{90}{\textbf{Wayne}}}\par\vspace*{0.08\textheight}
\vspace*{0.125\textheight}
\raisebox{0.5\height}{\rotatebox{90}{\textbf{Scott}}}
\end{minipage}
\hfill
\begin{minipage}[t]{0.46\textwidth} 
\centering
\textbf{Covering the Face Gesture}\par
\vspace{1mm} 

\makebox[0.33\linewidth][c]{\textbf{Source motion}}%
\makebox[0.33\linewidth][c]{\textbf{JMI result motion}}%
\makebox[0.33\linewidth][c]{\textbf{Base motion}}

\vspace{1mm}
\includegraphics[width=\linewidth,keepaspectratio]{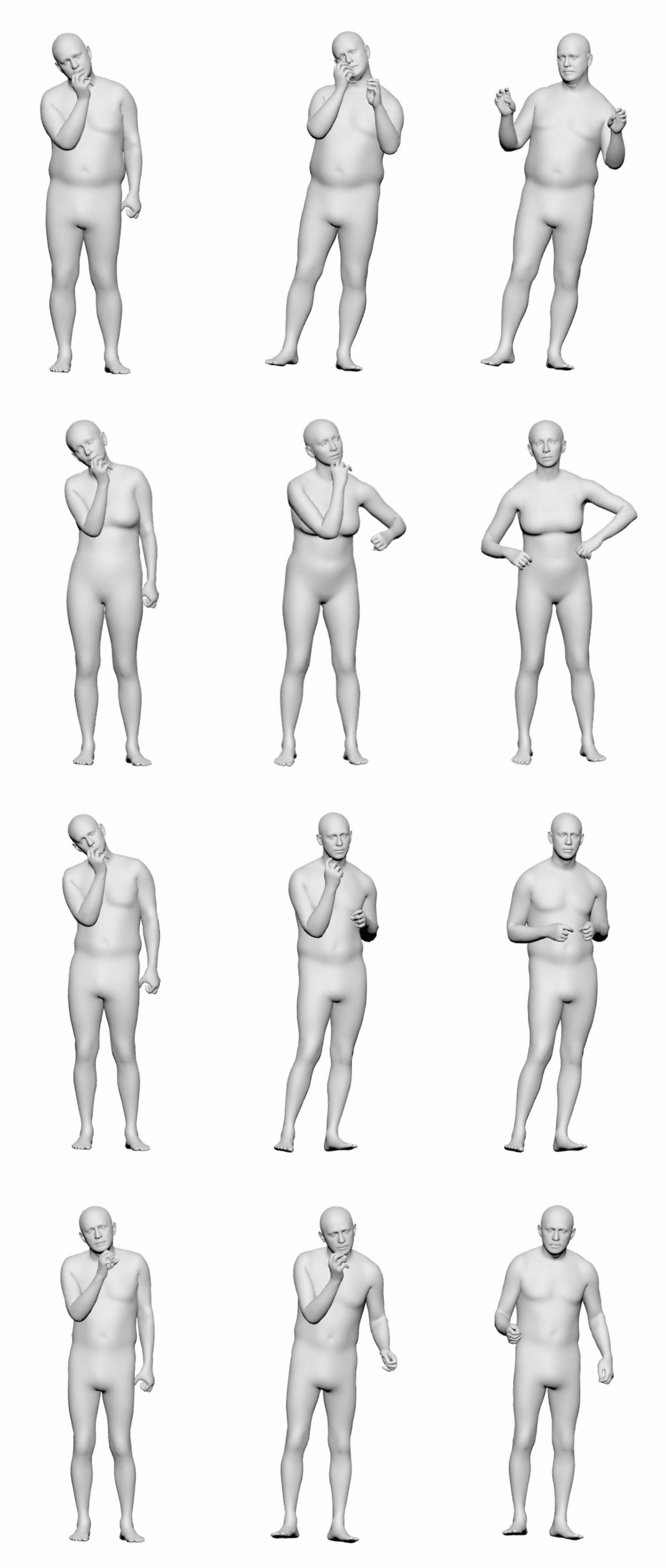}
\end{minipage}
\vrule
\begin{minipage}[t]{0.46\textwidth} 
\centering
\textbf{Thumbs Down Gesture}\par
\vspace{1mm} 

\makebox[0.33\linewidth][c]{\textbf{Source motion}}%
\makebox[0.33\linewidth][c]{\textbf{JMI result motion}}%
\makebox[0.33\linewidth][c]{\textbf{Base motion}}

\vspace{1mm}
\includegraphics[width=\linewidth,keepaspectratio]{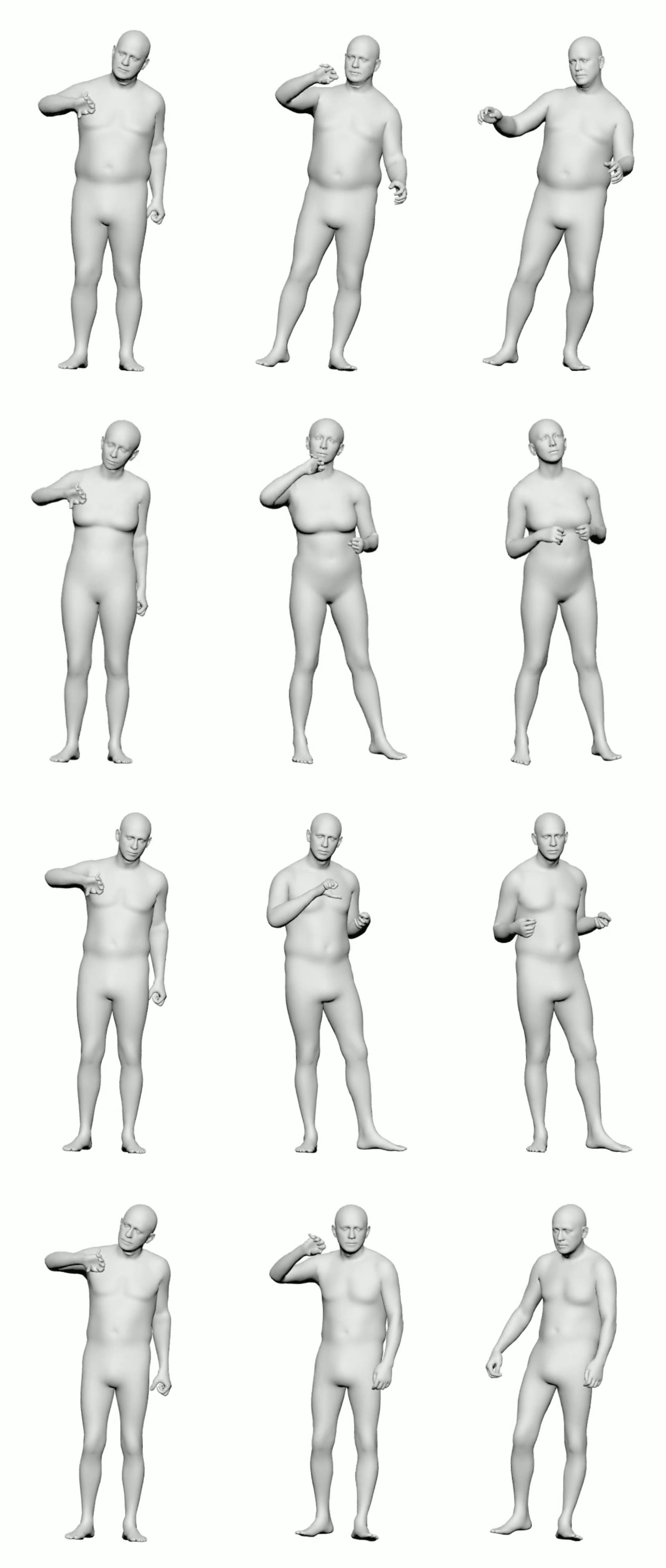}
\end{minipage}

\caption{\textbf{Cross-identity semantic gesture injection.} We show two semantic gestures, \textbf{Covering the Face (Hands Up)} (left) and \textbf{Thumbs Down} (right), applied to multiple target speakers (rows: Lawrence, Ayana, Wayne, Scott). For each gesture, columns compare the \textbf{source semantic motion}, our \textbf{JMI-injected result}, and the \textbf{base diffusion output} without injection. JMI enforces the intended semantic motion while preserving each speaker’s characteristic posture and motion style.}
\label{fig:qualitative2}
\end{figure*}

\clearpage

\section*{Acknowledgments}
This work was partially supported by the Horizon 2020 FET Proactive project GuestXR (\#101017884), the Israel Science Foundation (Grant no. 1427/25), and the Joint NSFC-ISF Research (Grant no. 3077/23).

\printbibliography  
\clearpage
\appendix

\appendix

\section{BVH to SMPL-X Retargeting}

Our method operates in the SMPL-X rotation space, while the SeG dataset, which provides semantic gesture clips, is in BVH format. To bridge this gap, we developed a robust pipeline to retarget all SeG motions to a unified 55-joint SMPL-X rig (body + hands + head). This section details the conversion process.
\begin{figure}[h]
    \centering
    \includegraphics[width=0.9\linewidth]{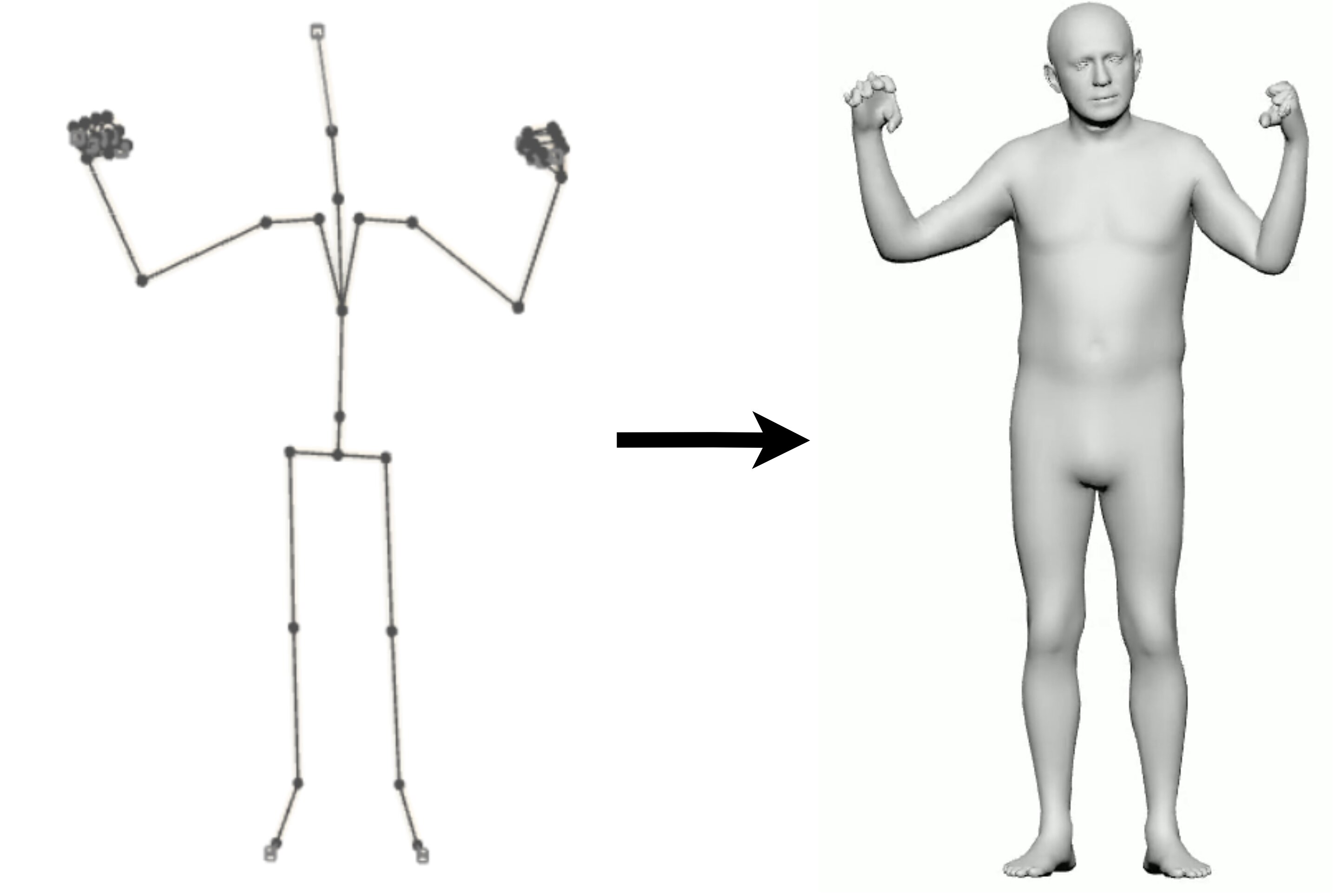} 
    \caption{Illustration of the BVH to SMPL-X retargeting process. On the left, a BVH skeleton (typically represented by sticks and joints) captures motion data. On the right, the same motion is retargeted onto an SMPL-X body model, demonstrating how the raw skeletal data is mapped onto a SMPL-X body representation for our framework.}
    \label{fig:bvh_to_smplx}
\end{figure}

\textbf{BVH Parsing and Joint Mapping.}
The initial step involves parsing each BVH file to extract the skeletal hierarchy, per-joint offsets, and local joint rotations. We then establish a deterministic mapping from the SeG skeleton to the SMPL-X kinematic tree. Key joints such as the pelvis/hips, spine, neck/head, shoulders, elbows, wrists, and fingers, as well as facial anchor joints, are meticulously mapped to their corresponding SMPL-X counterparts. Joints without a direct semantic match are either ignored or assigned to their closest kinematic neighbor within the SMPL-X structure. This process ensures that for each frame, a set of local BVH joint transforms is generated, maintaining structural consistency with the SMPL-X hierarchy.

\textbf{Scale and Root Alignment.}
BVH skeletons often exhibit variations in global scale and coordinate conventions. To standardize these, we normalize the scale by matching a reference body measurement (e.g., shoulder width or overall body height) between the BVH skeleton and a canonical SMPL-X template. A global scale factor is then uniformly applied to all joint offsets and root translations. The character is also recentered by positioning the SMPL-X pelvis at the origin at the first frame and eliminating any constant horizontal offset. Finally, the forward direction is standardized (e.g., aligning the character to face $+Z$) to ensure consistency across retargeted motions from different clips and datasets.

\textbf{Temporal Resampling and Audio Alignment.}
To align with our model setup, all BVH motions are resampled to 30 frames per second (FPS). Given the original frame times ${t_i}$ and target times ${\tilde{t}_k}$, we interpolate per-joint local rotations and root translations. Rotations are interpolated within their respective rotation spaces (e.g., via quaternion or matrix interpolation), while translations are linearly interpolated. This procedure generates temporally uniform SMPL-X pose trajectories that are precisely aligned with the audio and transcript timing used in our co-speech generation setting.

\textbf{IK Refinement and Jitter Suppression.}
Direct BVH retargeting can sometimes introduce undesirable artifacts. To mitigate this, we optionally apply a lightweight Inverse Kinematics (IK)-based refinement. This refinement enforces plausible contacts on selected joints and corrects large pose discrepancies that may arise from skeletal differences. Additionally, a mild temporal smoothing filter is applied to local joint rotations and root translations. This step effectively reduces high-frequency jitter while meticulously preserving the characteristic shape and timing of each gesture.

\textbf{Export to SMPL-X Rotation Space.}
The final step involves expressing each frame as SMPL-X pose parameters. For every mapped joint, its local rotation relative to the SMPL-X parent is computed within a canonical SMPL-X rest pose. These rotations are then stored in our 6D rotation representation (55 joints $\times$ 6D), alongside the global root translation trajectory. Crucially, we do not transfer shape-dependent or identity-specific parameters; the retargeting process is shape-agnostic, allowing semantic gestures to be universally applied to any SMPL-X speaker. The resulting $(T \times 55 \times 6)$ sequences are utilized both to populate our gesture bank and for evaluating semantic injection performance in the main paper.



\section{Quantitative Analysis of JMI Injection}
Table~\ref{tab:jmi_active_rest} validates the spatial precision of our Joint Motion Integration (JMI) mechanism. The results indicate that ``Active joints'' (those directly involved in the semantic action) exhibit a low distance to the target gesture, confirming successful semantic injection. Simultaneously, these active joints show a high deviation from the base motion, demonstrating that the semantic gesture has been effectively imposed. Conversely, ``Non-active joints'' maintain a low distance to the base motion (approximately 3–11 cm). This crucial finding demonstrates that the speaker's original style, posture, and rhythm are effectively preserved in body parts not involved in the semantic action, underscoring JMI's ability to selectively modify only the intended regions.

\begin{table}[t]
  \centering
  \caption{\textbf{Joint-space distances for semantic injection.}
  Mean Euclidean distance (cm) between our result and the base motion
  or SeG target, averaged over active vs.\ rest joints on the injection
  window. Lower is better.}
  \label{tab:jmi_active_rest}
  \begin{tabular}{lcc|cc}
    \toprule
    & \multicolumn{2}{c}{\textbf{Active joints}}
    & \multicolumn{2}{c}{\textbf{Non-active joints}} \\
    & \textbf{target}
    & \textbf{base}
    & \textbf{target}
    & \textbf{base} \\
    \midrule
    Scott     & 12.9 & 28.6 & 17.5 & 7.3 \\
    Wayne     & 14.5 & 28.1 &   20.7 &  6.0 \\
    Ayana     & 12.6 & 33.0 & 21.2 & 11.0 \\
    Sophie    & 15.8 & 29.7 &  14.4 & 3.1 \\
    Lawrence  & 11.8 & 26.7 &  18.4 & 7.3 \\
    \bottomrule
  \end{tabular}
\end{table}

\section{Perceptual Studies: Detailed Methodology}
\label{sec:sup-perceptual-details}

To evaluate the effectiveness of our method and its core mechanism, Joint Motion Integration (JMI), we conducted three distinct, anonymous perceptual studies. These were designed to measure the overall quality of co-speech gesture generation (Study I), the integration quality of semantic gestures (Study II), and the specific impact of the JMI module (Study III).

\subsection{General Protocol and Rendering}
To ensure participants focused exclusively on body motion, we normalized the visual presentation across all studies. 
\begin{itemize}
    \item \textbf{Rendering:} For Study I and III, animations were rendered using SMPL-X mesh body representations. For Study II (comparison against \textit{Semantic Gesticulator}), we utilized skeletal visualizations for both methods to ensure a fair comparison independent of rendering differences, as the baseline uses a BVH skeleton format.
    \item \textbf{Facial Features:} Facial expressions were specifically excluded to prevent distractors.
    \item \textbf{Randomization:} For every side-by-side comparison, the evaluated methods were anonymized, and their display order (left vs. right) was randomized to prevent presentation bias.
    \item \textbf{Audio:} Participants were encouraged to use headphones for optimal audio clarity and asked to answer using a computer rather than a mobile device.
\end{itemize}

\subsection{Participant Demographics}
We recruited a total pool of 54 unique participants across the three studies. The demographic profile was highly educated and technically literate:
\begin{itemize}
    \item \textbf{Background:} Over 90\% of participants were STEM professionals, including AI researchers, algorithm engineers, software developers, and Computer Science PhD candidates. The remaining participants represented diverse fields such as Medicine and Psychology. Notably, the pool included a body language expert and professional communication coaches, providing unique insights into gesture quality.
    \item \textbf{Age Profile:} The cohort consisted primarily of young adults and established professionals: 80\% aged 25--34, 13\% aged 35--44, and 7\% aged 45--64.
\end{itemize}

\subsection{Study I: Co-Speech Naturalness (Backbone)}
This study assessed the overall quality of co-speech gestures generated from input speech against four baselines: EMAGE, GestureLSM, SynTalker, and Ground Truth.

\textbf{Experimental Design.}
We recruited \textbf{32 participants} for this study. Each participant completed a total of \textbf{8 trials}. In each trial, they compared our method against one of the four baselines. Two distinct video samples were used for each baseline comparison to ensure robustness (4 baselines $\times$ 2 samples = 8 questions per participant).

\begin{figure}[h]
\centering
\includegraphics[width=\linewidth]{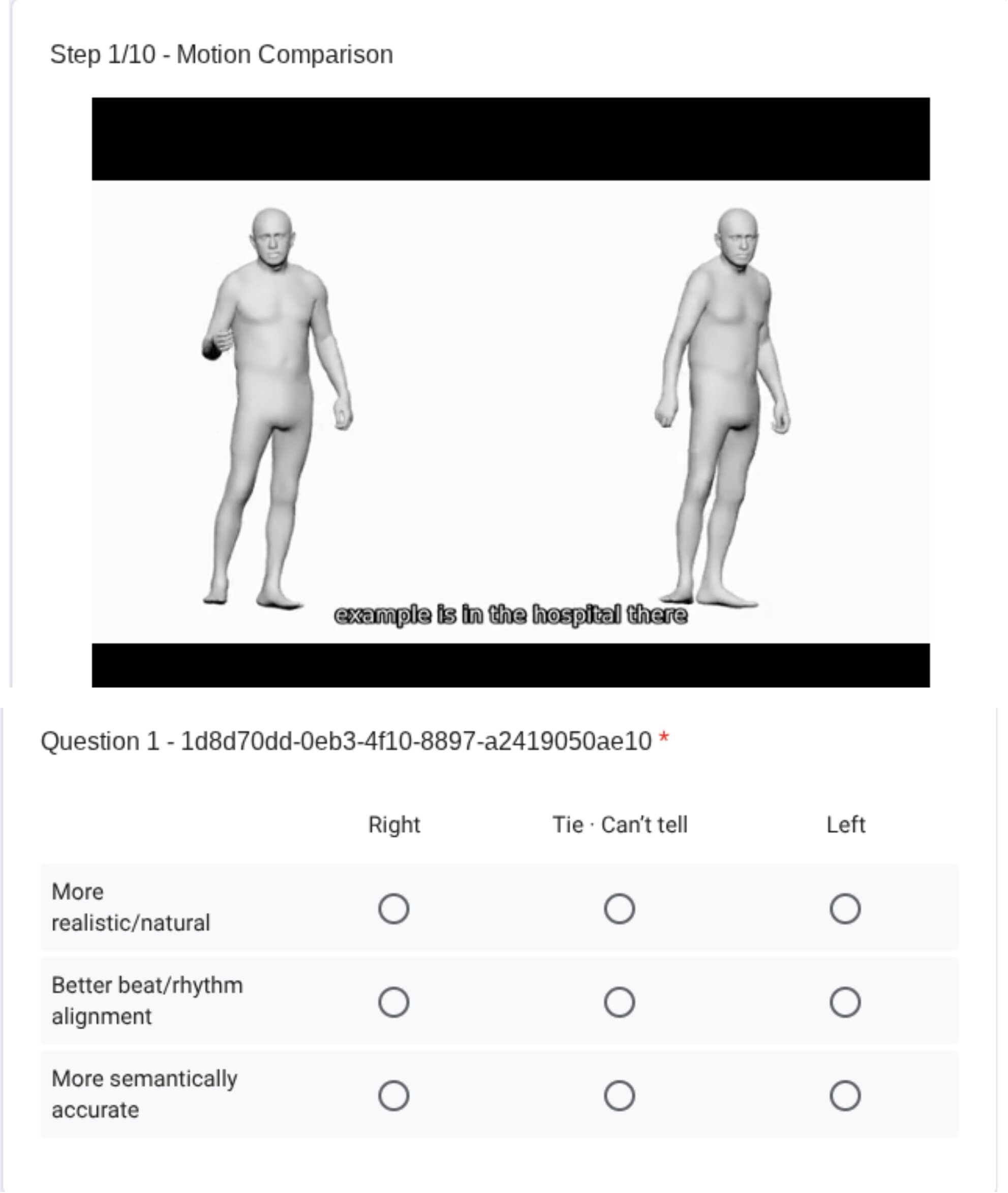}
\caption{Screenshot from Perceptual Study 1: Co-Speech Gesture Generation Quality. Participants compared two avatar animations side-by-side and evaluated their realism, sound-motion alignment, and semantic accuracy. This image depicts a practice pair used to familiarize participants with the task.}
\label{fig:perceptual_study1}
\end{figure}

\textbf{Task Description.} 
In each trial, participants viewed paired $\sim$7-second video clips side-by-side. They selected their preference (or a ``Tie'') based on three key criteria:
\begin{enumerate}
  \item \textbf{Realism / Naturalness:} ``Which video looks more realistic / natural?''
  \item \textbf{Audio-Motion Alignment:} ``Which video is better aligned with the audio beat/rhythm?''
  \item \textbf{Semantic Accuracy:} ``Which video’s gestures better match the meaning of the speech?''
\end{enumerate}
A screenshot of the interface is shown in Figure~\ref{fig:perceptual_study1}.

\subsection{Study II: Semantic Injection Quality}
This study evaluated our Joint Motion Integration (JMI) method by examining how effectively external ``target gestures'' were injected into a continuous base animation compared to the \textit{Semantic Gesticulator} baseline.

\textbf{Experimental Design.}
We recruited \textbf{22 participants} for this study. Each participant evaluated \textbf{4 distinct semantic injection scenarios}. This resulted in a total of 88 evaluations ($22 \times 4$), providing a statistically significant sample size for analysis.

\begin{figure}[htbp]
  \centering
  
  \subfloat[Evaluation Interface]{
    \includegraphics[width=\linewidth]{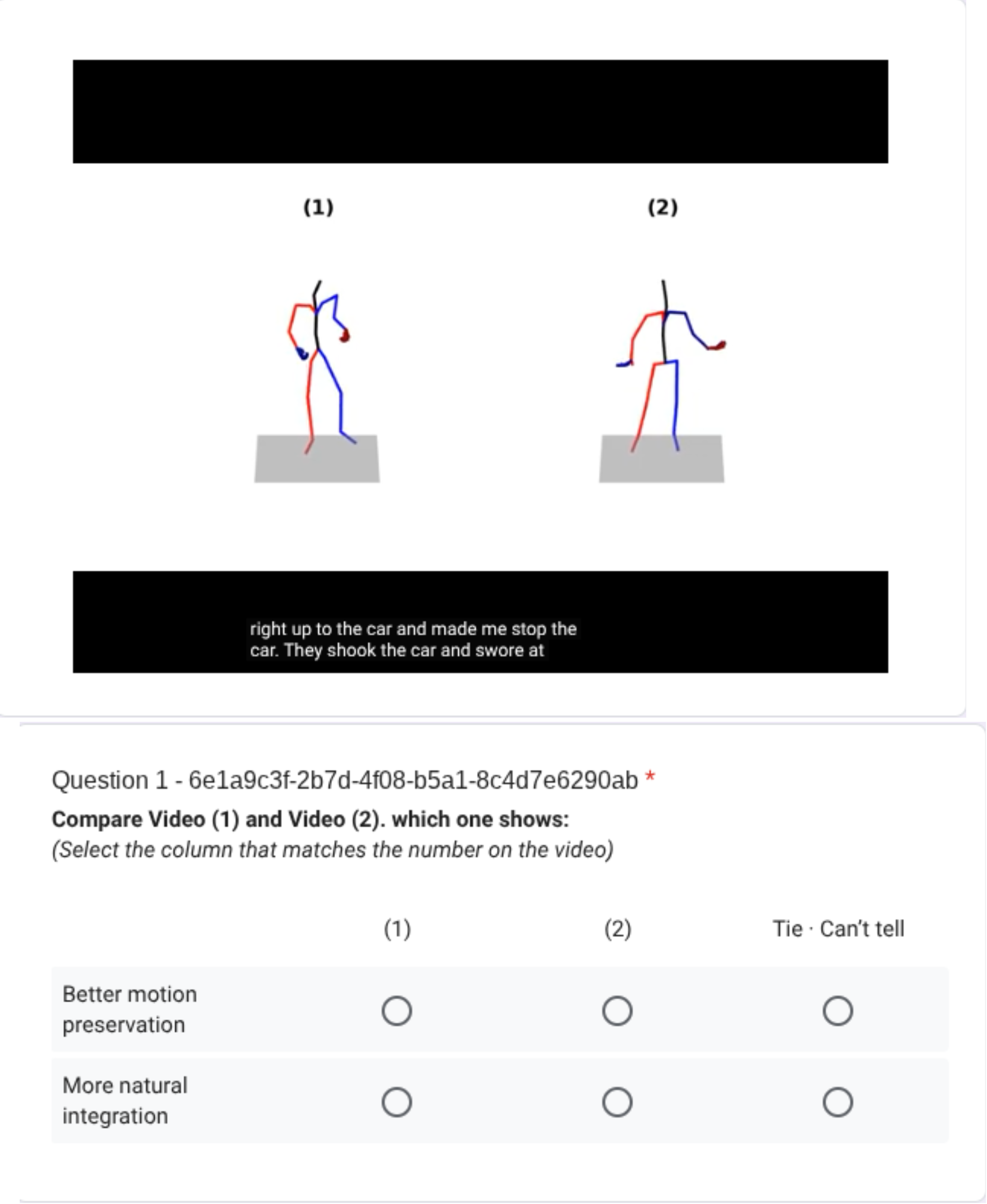}
  }
  
  \vspace{1em} 
  
  \subfloat[Source Motion Reference]{
    \includegraphics[width=\linewidth]{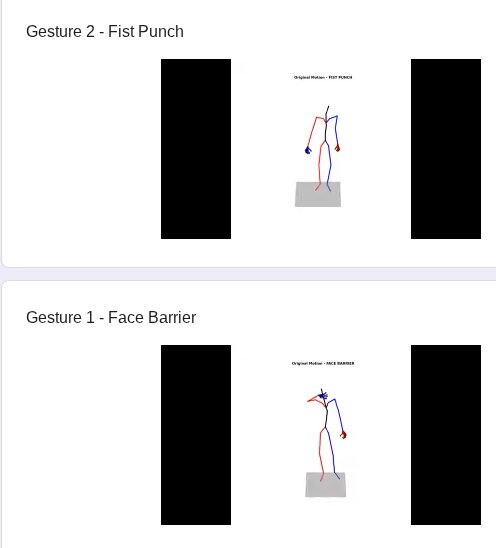}
  }
  
\caption{Screenshot from Perceptual Study 2: Semantic Gesture Integration. This image illustrates how target gestures (e.g., ``Fist Punch,'' ``Face Barrier'') were presented using skeletal motion rigs as ground truth references, before participants evaluated their integration into full animations.}
\label{fig:perceptual_study2}
\end{figure}

\textbf{Task Description.} 
Participants were first shown a reference ``source'' motion from the SeG dataset (containing 2--3 semantic tags), serving as the ground truth. They then viewed generated outputs for the same utterance.
Figure~\ref{fig:perceptual_study2} illustrates the presentation format.

\textbf{Evaluation Criteria.}
Participants compared our method against the baseline using two specific prompts:
\begin{enumerate}
  \item \textbf{Motion Preservation (Fidelity):} ``Which video better preserves the specific movements from the Original Motion?''
  \item \textbf{Natural Integration (Seamlessness):} ``Which video blends the gestures more naturally into the rest of the animation?''
\end{enumerate}

\subsection{Study III: Perceived Impact of JMI (Ablation)}
To validate JMI's critical impact on perceived quality, we compared our Full Model (with JMI-injected gestures) against the Diffusion Backbone alone (without explicit semantic injection).

\textbf{Experimental Design.}
This study was conducted as an extension of Study I using the same pool of \textbf{32 participants}. Following the main baseline comparisons, participants completed \textbf{2 additional trials} comparing the Full Model against the Backbone. This resulted in a total of 64 pairwise comparisons.

\textbf{Task Description.}
To ensure consistency, this study utilized the \textbf{identical interface and protocol as Study I}. Participants evaluated short videos containing 2--4 gesture injection points and answered the same three questions regarding Realism, Alignment, and Semantic Accuracy.

\section{\ar{Isolating the Contribution of the
Global CLIP Prior}}
\ar{Our main-text ablation removes CLIP and FiLM jointly
(“Ours w/o CLIP\&FiLM”). To isolate the global CLIP embedding’s specific contribution, we run two further experiments.}

\textbf{\ar{Inference-time prompt ablation.}}
\ar{Keeping the trained model fixed, we vary only the text prompt supplied at inference and measure its effect on distribution realism. Using the speaker transcript yields our best FGD (4.72). Dropping the text condition entirely (“None”) degrades it to 5.37, and supplying a mismatched prompt (“I am tired”) degrades it sharply to 10.88. This shows the global prior meaningfully shapes the output distribution, and that a semantically wrong prompt is actively harmful rather than merely uninformative.}

\textbf{\ar{CLIP-only conditioning.}}
\ar{We additionally evaluate a variant conditioned exclusively on the global CLIP embedding, removing the local audio and transcript token streams. This collapses performance (FGD 24.90), confirming that the global prior alone cannot drive rhythmically accurate co-speech generation.

Together these results indicate that the CLIP prior serves mainly as a supplementary global prior, while the local audio/text streams primarily drive rhythmic synchronization and stylistic realism}

\section{\ar{LLM Tagging Prompt and Examples}}
\ar{Semantic gesture tagging is performed with a single zero-shot GPT-4 call, followed by manual post-filtering. Our prompt is adapted from the gesture-annotation prompt of Semantic Gesticulator \cite{zhang2024semantic}. The model receives a task description, the utterance transcript, and semantic gesture dictionary (each entry: gesture name, a short textual description, Contextual Meaning, and an example). It returns the transcript with inline gesture annotations attached
to the relevant words.

\textbf{Tagging Prompt~\cite{zhang2024semantic}.}

\begin{quote}
\ttfamily
\small
``mark the given transcripts with gestures. Here are the
gestures you can use: \\
3100  ARM  FLEX  The arm is (or arms are) bent to produce maximum bulging of the muscles.. 0210 FOREHEAD SALUTE.. 3010  ARM RAISE HIGH-LEVEL ...\\
Input:\\
I think it's safe to say there's never been a moment like this one in
all of history.\\
Output:\\
I (3332 TEMPLE TOUCH) think it's safe to say there's never been a moment like this one in all of (2030 ARMS DESCEND) history.''

\end{quote}

Example input transcript:
\begin{quote}
\ttfamily
\small
``When I was at university which is almost more than 10 years ago the community security was not good at the time and we were often more than not in dangerous situations one time I was walking home from work and waiting at the bus stop I heard the sound of gunfire  from across the road police cars came around and the police shot at people across the street I was so scared the main reason was you have no place to hide you're at a bus station there's no place to go there was another time that I was in Los Angeles driving I took the wrong road it went down a small local road I'm not exaggerating but here on both sides of the road we're filled with homeless people those people were moving around the road and I was very nervous they ignored the traffic rules and came right up to my car and made me stop my car they shook the car and swore at me I thought to myself this is horrible it's a few days ago someone was shot dead and ..''
\end{quote}
Example tagging output:

\begin{quote}
\ttfamily
\small
``when I was at university which is almost more than 10 (1232 FOREFINGER SPIRAL) years ago the community security was not good (2121 THUMB DOWN) at the time and we were often more than not in dangerous (1022 HANDS BEAST) situations one (1231 FOREFINGER RAISE-ONE) time I was walking home from work and waiting (0212 WRIST CHECK-TIME) at the bus stop I heard the sound of gunfire (1100 ARMS EXPLODE) from across the road police cars came around and the police shot (0121 TEMPLE 'SHOOT') at people across the street I was so scared (0131 FACE COVER) the main reason was you have no place to hide you're at a bus station there's no place to go (0322 SHOULDERS SHRUG) there was another time that I was in Los Angeles driving (1302 HANDS STEERING) I took the wrong road it went down a small local road I'm not exaggerating but here on both sides of the road we're filled (3032 ARMS ENCOMPASS) with homeless people those people were moving around the road and I was very nervous they ignored the traffic rules and came right up to my car and made me stop (1133 FIST PUNCH) my car they shook the car and swore at me I thought to myself this is horrible (0221 FACE BARRIER) it's a few days ago someone was shot dead and ..''
\end{quote}
We note that this LLM step is optional: annotations may equally be supplied manually, and our injection pipeline is agnostic to how they are produced. Precise gesture descriptions in the dictionary are important to tagging accuracy; deeper exploration of fully-automated semantic retrieval is left to future work.}

\section{\ar{Per-Step JMI Intervention --- Implementation Detail}}
\label{app:jmi-detail}
\ar{This appendix expands the per-step joint-space intervention of Section~3.3 of the main paper. During inference, at each diffusion step $t$, the backbone denoiser predicts a clean motion $\hat{x}_0 = f_\theta(x_t, t, \mathrm{cond})$ for every generation window. For a given window indexed by $w$, we first identify all tagged semantic gestures in $\mathcal{I}$ that intersect its timeframe. For each intersecting specification, we modify the initial prediction $\hat{\mathbf{x}}_0$ through the following steps:

\textbf{Projection and cropping.}
For a generation window with global start time $T_w$, we project the global semantic interval $[s^*_k, e^*_k]$ onto local window indices $[s^w_k, e^w_k]$ satisfying
\begin{equation}
  s^*_k \;\le\; T_w + s^w_k \;<\; T_w + e^w_k \;\le\; e^*_k .
\end{equation}
We then extract the corresponding feature slice
$G^m_{\mathrm{crop}} \in \mathbb{R}^{6J \times L}$ from the gesture sequence, where $L = e^w_k - s^w_k$ is the crop length.

\textbf{Mixing matrix construction.}
We build a 1D temporal profile $\tau \in [0,1]^{L}$ with a trapezoidal shape:
linear ramps of length $B_{\mathrm{blend}}$ at the two boundaries and unity in the interior. The spatiotemporal mixing matrix is obtained by broadcasting this profile across the active joints via an outer product,
\begin{equation}
  \Phi \;=\; \lambda_{\mathrm{inj}} \left( M^m \otimes \tau \right)
  \;\in\; [0,1]^{6J \times L},
\end{equation}
where $\lambda_{\mathrm{inj}} \in [0,1]$ is the global injection strength and $M^m \in \{0,1\}^{6J}$ is the active-joint mask.

\textbf{Composition.}
The prediction is updated by convex blending over the temporal slice $[s^w_k : e^w_k]$:
\begin{equation}
  \hat{x}^{\,\mathrm{edit}}_0\!\left[:6J, s^w_k\!:\!e^w_k\right]
  \;\leftarrow\;
  (1 - \Phi) \odot \hat{x}_0\!\left[:6J, s^w_k\!:\!e^w_k\right]
  \;+\; \Phi \odot G^m_{\mathrm{crop}} .
\end{equation}
The modified prediction $\hat{x}^{\,\mathrm{edit}}_0$ then replaces $\hat{x}_0$ in the standard DDIM posterior sampling step to obtain $\hat{x}^{\,\mathrm{edit}}_{t-1}$. Because $\Phi_{ij} \approx 0$ off the active joints, those joints retain the backbone's own prediction at the current step, preserving the speaker's style and natural motion, while the active joints are pulled toward the injected semantic gesture.}

\end{document}